\pdfoutput=1
\documentclass[11pt]{article}

\usepackage[final]{acl}

\usepackage{times}
\usepackage{latexsym}

\usepackage[T1]{fontenc}
\usepackage[utf8]{inputenc}

\usepackage{microtype}

\usepackage{inconsolata}

\usepackage{graphicx}

\usepackage{multirow}
\usepackage{booktabs}
\usepackage{array}
\usepackage{subcaption}
\usepackage{bm}
\usepackage{amsfonts}
\usepackage{amsmath}
\DeclareMathOperator{\argmax}{arg\,max}
\usepackage{tcolorbox}
\usepackage{colortbl}
\usepackage{xcolor}
\definecolor{SoftBlue}{HTML}{4682B4}
\definecolor{SoftRed}{HTML}{D15D5D}

\newcommand{\red}[1]{\textcolor{red!80}{#1}}
\newcommand{\green}[1]{\textcolor{green!80}{#1}}
\newcommand{\highlight}[1]{\textcolor{SoftBlue}{\textbf{\textit{#1}}}}
\newtheorem{theorem}{Theorem}{}

\newtheorem{lemma}[theorem]{Lemma}

\usepackage{helvet}
\usepackage{algorithm}
\usepackage{algorithmic}

\title{Prompt Candidates, then Distill: A Teacher-Student Framework for LLM-driven Data Annotation}

\author{First Author \\
  Affiliation / Address line 1 \\
  Affiliation / Address line 2 \\
  Affiliation / Address line 3 \\
  \texttt{email@domain} \\\And
  Second Author \\
  Affiliation / Address line 1 \\
  Affiliation / Address line 2 \\
  Affiliation / Address line 3 \\
  \texttt{email@domain} \\}

\author{
 \textbf{Mingxuan Xia\textsuperscript{1}},
 \textbf{Haobo Wang\textsuperscript{1}\thanks{$\;$ Corresponding author.}},
 \textbf{Yixuan Li\textsuperscript{2}},
 \textbf{Zewei Yu\textsuperscript{1}},
\\
 \textbf{Jindong Wang\textsuperscript{3}},
 \textbf{Junbo Zhao\textsuperscript{1}},
 \textbf{Runze Wu\textsuperscript{4}}
\\
 \textsuperscript{1}Zhejiang University
 \textsuperscript{2}University of Wisconsin Madison \\
 \textsuperscript{3}William \& Mary
 \textsuperscript{4}NetEase Fuxi AI Lab \\
 \texttt{\{xiamingxuan,wanghaobo,22451274,j.zhao\}@zju.edu.cn}\\
 \texttt{sharonli@cs.wisc.edu,jwang80@wm.edu,wurunze1@corp.netease.com}\\
}

\begin{document}
\maketitle
\begin{abstract}
% Recently, Large Language Models (LLMs) have demonstrated great potential for data annotation, particularly in active learning, serving as a human-free active annotator which minimizes the annotation cost to train a model on specific NLP tasks. However, due to the notorious hallucination problem, existing LLM-driven annotation methods inevitably generate completely wrong answers, leading to limited performance of the trained model. To alleviate this problem, we propose CanDist, a candidate annotation-enhanced LLM-driven active learning framework, which prompts LLM to generate candidate annotations and trains a Small Language Model (SLM) on those candidates. Specifically, our motivation comes from Ambiguity Aversion where people tend to answer more conservatively, i.e., providing multiple candidate answers, and we statistically illustrate that candidate annotation retains more valuable information than single annotation. The SLM is trained with a distriution refinery mechanism to identify the ground-truth labels embodied in candidate annotations. We further present a rigorous justification that training on candidate annotations enjoys better theoretical guarantees than training on single ones. Experiments on six text classification tasks demonstrate the effectiveness of CanDist.
Recently, Large Language Models (LLMs) have demonstrated significant potential for data annotation, markedly reducing the labor costs associated with downstream applications. However, existing methods mostly adopt an aggressive strategy by prompting LLM to determine a single gold label for each unlabeled sample. Due to the inherent uncertainty within LLMs, they often produce incorrect labels for difficult samples, severely compromising the data quality for downstream applications. 
Motivated by ambiguity aversion in human behaviors, we propose a novel candidate annotation paradigm wherein large language models are encouraged to output all possible labels when incurring uncertainty. 
To ensure unique labels are provided for downstream tasks, we develop a teacher-student framework CanDist that distills candidate annotations with a Small Language Model (SLM).
We further provide a rigorous justification demonstrating that distilling candidate annotations from the teacher LLM offers superior theoretical guarantees compared to directly using single annotations. Extensive experiments across six text classification tasks validate the effectiveness of our proposed method. The source code is available at \url{https://github.com/MingxuanXia/CanDist}.
\end{abstract}

\section{Introduction}

Various NLP tasks require collecting high-quality labeled data for model training (e.g. text classification \citep{DBLP:journals/information/KowsariMHMBB19}, named entity recognition \citep{DBLP:journals/tkde/LiSHL22}, and sentiment analysis \citep{DBLP:journals/air/WankhadeRK22}), which typically involves human experts meticulously providing high-quality target labels, a process that is notoriously time-consuming and labor-intensive.
With the development of Large Language Models \citep{DBLP:journals/corr/abs-2303-08774, DBLP:journals/corr/abs-2312-11805, DBLP:journals/corr/abs-2407-21783}, LLM-driven automatic data annotation approaches have been proposed \citep{DBLP:journals/corr/abs-2303-15056, DBLP:conf/emnlp/TanLWBJBKL0024, DBLP:conf/acl/LongWXZDCW24}, relieving the burden of the cost-prohibitive human annotation.

\begin{figure}[!t]
    \centering
    \includegraphics[width=1.0\linewidth]{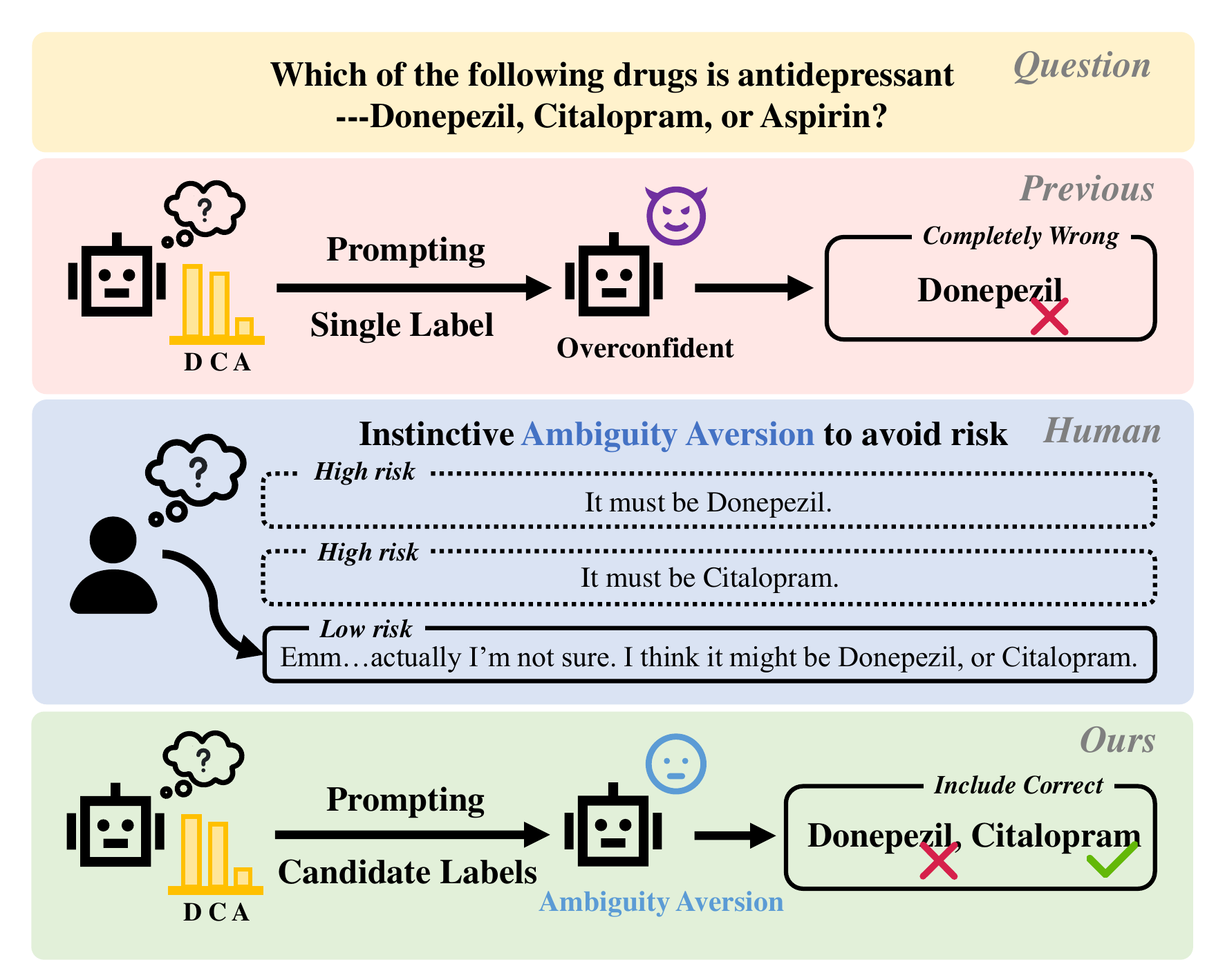}
    \caption{When facing uncertainty, humans instinctively behave ambiguity aversion to avoid risk, which motivated us to prompt LLM for candidate annotations (multiple possible answers), increasing the likelihood of providing the correct labels.}
    \label{fig:aa}
\end{figure}

Although LLMs excel at general language understanding and generation, their knowledge of downstream tasks remains limited \citep{DBLP:conf/acl/LiCRCZNW24}. As a result, LLMs may be uncertain about some samples during annotation. Nevertheless, existing LLM-driven annotation methods prompt LLMs with \textbf{single annotation}, which forces the model to assign a specific label to each unlabeled sample---even when it is unsure. This often leads to completely wrong annotations, which is not only a waste of computational resources but also affects downstream training \citep{DBLP:conf/acl-insights/ZhuHZAK22}. Moreover, it necessitates further error localization and re-labeling, which is both costly and time-consuming. This raises a critical question: \textit{Can we induce LLMs to provide a more valuable annotation rather than a completely wrong label when they are uncertain?}

\begin{figure}[!t]
    \centering
    \includegraphics[width=0.99\linewidth]{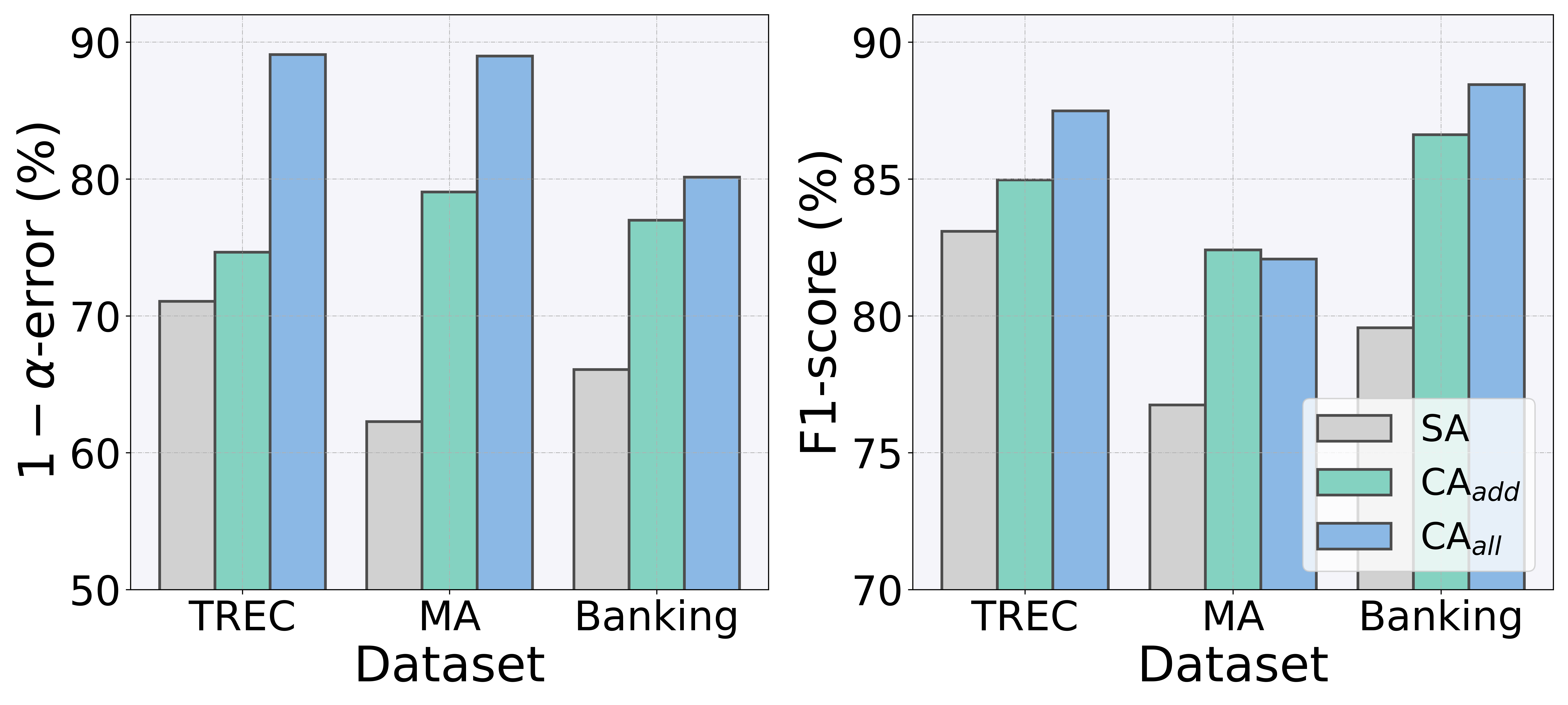}
    \caption{Comparison of $1-\alpha$-error and F1-score between single annotations (SA) and candidate annotations (CA) by GPT-3.5. Higher metric values indicate better results. See section \ref{subsec:generation} for details.}
    \label{fig:gpt_assessment}
\end{figure}

To answer this question, we first draw an analogy to human behavior---when faced with uncertainty, humans often behave conservatively instead of being overconfident---an instinctive psychological phenomenon known as \textit{Ambiguity Aversion} \citep{fox1995ambiguity, maccheroni2006ambiguity}. This behavior helps people mitigate severe risks and ensures the lower bound of the gains.
% To answer this question, we first draw an analogy to human behavior---when faced with uncertainty, humans often behave conservatively instead of being overconfident. For instance, in healthcare, patients tend to avoid treatments that involve uncertainty; In legal judgments, judges are inclined to avoid ruling on cases with uncertain evidence; In investment, people tend to steer clear of investments with unclear probabilities of returns.
% This is an instinctive psychological phenomenon known as \textit{Ambiguity Aversion} \citep{fox1995ambiguity, maccheroni2006ambiguity}, which helps people mitigate severe risks and ensures the lower bound of the gains.
Motivated by this, we propose to induce LLMs to exhibit ambiguity aversion during annotation, by prompting them to provide multiple possible labels for each unlabeled sample, i.e., \textbf{candidate annotations}. As shown in Figure \ref{fig:aa}, although the LLM may fail to provide a correct answer with a single label, answering with candidate labels successfully includes the correct one. We further demonstrate in Figure \ref{fig:gpt_assessment} that, on a macro level, candidate annotations are more likely to cover correct labels (higher $1-\alpha$-error) and retain more value (higher F1-score) than single annotations. Note that, unlike methods such as Self-Consistency \citep{DBLP:conf/iclr/0002WSLCNCZ23}, prompting candidates is asking for the inherent uncertainty rather than randomness, see Table \ref{tab:ab_generate} for detailed discussion.

% To address this issue, we show that Small Language Models (SLMs) are efficient and effective learners in identifying the correct label from multiple candidates. 
Despite its great potential, however, candidate annotations cannot be directly applied to downstream tasks, as they require one specific label for each sample. To address this issue, we draw inspiration from knowledge distillation \citep{DBLP:journals/corr/HintonVD15} where the student model is distilled from the teacher's output distribution and exhibits better generalization on downstream tasks \citep{DBLP:conf/icml/PhuongL19}, and propose a teacher-student framework called \textbf{CanDist} that distills high-quality knowledge from the teacher LLM's candidate annotations to a student Small Language Model (SLM) to achieve data annotation. Specifically, we introduce a distribution refinery (DR) mechanism during distillation that dynamically adjusts the training target based on SLM's predictions, where correct labels gradually emerge from those false positive ones.
Theoretically, we justify that \textit{distilling from candidate annotations from the teacher LLM offers superior theoretical guarantees than directly using the single annotations from the teacher LLM}.
Empirically, we evaluate CanDist on six text classification tasks, where CanDist achieves state-of-the-art among various LLM and SLM baselines.

\section{Related Work}

\subsection{LLM for Data Annotation}
LLM-driven data annotation has been applied in various NLP tasks, such as text classification \citep{DBLP:journals/corr/abs-2303-15056}, relation extraction \citep{DBLP:conf/acl/DingQLCLJB23}, named entity recognition \citep{DBLP:journals/corr/abs-2402-14568}, question answering \citep{DBLP:conf/naacl/HeLGJZLJYDC24}, semantic parsing \citep{DBLP:conf/emnlp/ShinLTCRPPKED21}, and multilingual text generation \citep{DBLP:conf/eacl/ChoiLJK24}.
Advanced approaches adopt techniques like in-context learning \citep{DBLP:conf/nips/BrownMRSKDNSSAA20, DBLP:conf/emnlp/Xiao0ZWLCW23, DBLP:journals/corr/abs-2409-12425}, chain-of-thought prompting \citep{DBLP:conf/nips/Wei0SBIXCLZ22, DBLP:conf/naacl/HeLGJZLJYDC24, DBLP:conf/acl/YuanCZJ24}, and collaboration with fine-tuned SLMs \citep{DBLP:conf/emnlp/Xiao0ZWLCW23, DBLP:conf/acl/XuXWLZM24, DBLP:conf/iclr/YangZYBW0X00C024} to boost LLM's zero-shot performance for annotations.  

However, these approaches limit LLMs to provide single annotations, which inevitably introduce completely wrong labels. In contrast, we investigate a more conservative strategy by prompting LLMs for candidate annotations, which offers greater value. Besides, while FreeAL \citep{DBLP:conf/emnlp/Xiao0ZWLCW23}, the pioneering work of SLM-collaborated annotation, has demonstrated the effectiveness of distilling the SLM from LLM's single annotations, we propose that distilling from candidate annotations yields superior results and we rigorously provide its theoretical guarantees.

% Another line of work introduce SLM to assist LLM in data annotation , where fine-tuned SLMs can outperform LLMs on downstream tasks \citep{DBLP:conf/ijcnlp/BangCLDSWLJYCDXF23}. The pioneering work FreeAL \citep{DBLP:conf/emnlp/Xiao0ZWLCW23} proposes a collaborative framework where an SLM is used to distill high-quality task-related knowledge from the single weak annotations provided by the LLM.
% and in return feedback valuable few-shot examples to LLMs for label refinery.

\begin{table*}[!t]
    \caption{Key prompts of prompting single (SA) and candidate ($\text{CA}_\text{add}$ and $\text{CA}_\text{all}$) annotations on the TREC dataset.}
    \label{tab:prompts}
    \centering
    \small
    \renewcommand\arraystretch{1.2}
    \scalebox{1.0}{
    \begin{tabular}{p{1.5cm}p{13.2cm}}
    \toprule
        \textbf{Strategy} & \textbf{Prompt} \\ 
    \midrule
        SA & \parbox{13.2cm}{\textsf{Given a question: $\dots$ What does the question ask about? Please identify the question \highlight{into one} of the following types: \textbf{Abbreviation}; \textbf{Description and abstract concepts}; \textbf{Entities}; \textbf{Human beings}; \textbf{Locations}; \textbf{Numeric values}.}} \\ 
    \midrule
        $\text{CA}_\text{add}$ & \parbox{13.2cm}{\textsf{Given a question: $\dots$ What does the question ask about? Please identify the question into one of the following types: \textbf{Abbreviation}; \textbf{Description and abstract concepts}; \textbf{Entities}; \textbf{Human beings}; \textbf{Locations}; \textbf{Numeric values}. \highlight{If you are unsure about your answer, please include other potential choices.}}} \\
    \midrule
        $\text{CA}_\text{all}$ &  \parbox{13.2cm}{\textsf{Given a question: $\dots$ What does the question ask about? Please identify the question \highlight{with all possible choices} of the following types: \textbf{Abbreviation}; \textbf{Description and abstract concepts}; \textbf{Entities}; \textbf{Human beings}; \textbf{Locations}; \textbf{Numeric values}.}} \\ 
    \bottomrule
    \end{tabular}}
\end{table*}

\subsection{Generate and Aggregate Multiple Answers with LLM}
Recently, solving NLP tasks by generating multiple diverse answers using LLMs and then aggregating them to extract their essences has been increasingly popular.
Sampling-based strategy first samples a diverse set of reasoning paths during LLM decoding, and then integrate them through methods such as trained ranking models \citep{DBLP:journals/corr/abs-2110-14168, DBLP:conf/emnlp/ShenYLSJ0021, DBLP:journals/corr/abs-2201-08239}, majority voting \citep{DBLP:conf/iclr/0002WSLCNCZ23, DBLP:conf/iclr/FuPSCK23, li2022making}, LLMs \citep{DBLP:journals/corr/abs-2311-17311, DBLP:conf/emnlp/WengZX0HLSLZ23, DBLP:conf/acl/ZhangPTZJSMM24}, or human feedback \citep{DBLP:conf/emnlp/Li24}.
Ensemble-based methods generate multiple answers by gathering outputs from different prompt designs, such as different prompt formats \citep{DBLP:conf/emnlp/ZhouHMBN22, DBLP:journals/corr/abs-2310-03094, DBLP:conf/acl/ZhangSWPWZ024} or different permutations of few-shot examples \citep{DBLP:conf/icml/ZhaoWFK021, DBLP:conf/acl/LuBM0S22, DBLP:journals/corr/abs-2203-05115}.
Additionally, a few approaches propose to directly prompt candidates, in the applications of model calibration \citep{DBLP:conf/emnlp/TianMZSRYFM23, DBLP:conf/iclr/XiongHLLFHH24} and open-domain QA \citep{DBLP:conf/iclr/KimNMP0S0S24}.

However, sampling and ensemble-based methods rely on the randomness of LLMs, making them costly and inefficient in providing enough valuable annotations compared to prompting candidates. Moreover, this paper proposes a novel aggregation strategy that leverages an SLM to distill high-quality annotations from the multiple labels provided by the LLM.

% \subsection{Active Learning}
% Active learning \citep{DBLP:conf/icimcs/XuSZ13, DBLP:conf/emnlp/ZhangSH22a} is a prevailing paradigm where the model selectively queries the most informative data points to improve its performance with minimal labeled data. According to the querying strategy, traditional active learning methods can be categorized into uncertainty-based methods \citep{DBLP:conf/emnlp/PrabhuDS19, DBLP:conf/emnlp/MargatinaVBA21} and diversity-based methods \citep{DBLP:conf/iclr/SenerS18, DBLP:conf/iclr/AshZK0A20}. While traditional active learning methods still require costly human annotations, LLM-driven active learning methods \citep{DBLP:conf/emnlp/Xiao0ZWLCW23, DBLP:conf/acl/YuanCZJ24, DBLP:conf/naacl/LiangL0L024} have been recently proposed, offering an exciting solution with minimized annotation cost for NLP tasks. The pioneering work FreeAL \citep{DBLP:conf/emnlp/Xiao0ZWLCW23} proposes a collaborative framework where an SLM is leveraged to distill the high-quality task-related knowledge from the single weak annotation by LLMs and in return feedback valuable few-shot examples to LLMs for label refinery. Following this line of work, we revolutionize LLM-driven active learning with candidate annotations which provide more ground-truth labels and enjoy a better theoretical guarantee.

\section{Proposed Method}

\subsection{Preliminaries}
In this paper, we consider the task of text classification, where an \textbf{unsupervised} dataset $\mathcal{D}=\{\bm{x}_i\}_{i=1}^n$ with $n$ samples is provided. Given the label space $\mathcal{Y}=\{1, \dots, C\}$ with corresponding semantic meanings, each sample $\bm{x} \in \mathcal{X}$ is associated with a ground-truth label $y \in \mathcal{Y}$, which is inaccessible. In LLM-driven data annotation, an LLM $\mathcal{T}$ serves as the annotator, providing labels for the unlabeled samples in $\mathcal{D}$.
Most existing methods prompt LLMs to provide a \textbf{Single Annotation} (SA), i.e., a specific label $\tilde{y}_i\in\mathcal{Y}$ for each $\bm{x}_i$.

% aiming to label the unlabeled samples as close as possible to their ground-truth labels. Specifically, we formalize the data annotation process of LLM by defining a natural language template $T(\cdot)$ which queries the LLM a \textit{single} label for each unlabeled sample based on task-related information, and a verbalizer $V(\cdot)$ which maps each class label in $\{1, \dots, C\}$ to a pre-defined class token in the prompt. Formally, given a sample $x$, the prediction given by LLM is formalized as:
% \begin{equation}
%     \underset{y \in \mathcal{Y}}{\argmax} ~ P\left(V\left(y\right) | \{T(x_i^{d}, \tilde{y}_i^{d})\}_{i=1}^m, T\left(x\right)\right)
% \end{equation}
% where $\{x_i^d, \tilde{y}_i^d\}_{i=1}^m$ are demonstrations to conduct few-shot in-context learning (ICL) \cite{DBLP:conf/nips/BrownMRSKDNSSAA20} when supervised data are available. We refer to this process as zero/few-shot annotation.

\subsection{Prompt Candidate Annotations by LLM}\label{subsec:generation}
% Despite the promise of zero/few-shot learning, LLM suffers from the inherent hallucination problem which could generate factual errors and unfaithful answers, leading to incorrect labels during data annotation \citep{DBLP:journals/csur/JiLFYSXIBMF23, DBLP:conf/iclr/MundlerHJV24}. In CanDist, we propose a novel data annotation paradigm that queries the LLM a \textit{candidate} set of labels for each sample instead of a single one, where the correct label is more likely to be included, having more opportunities to be exploited and utilized. Our motivation comes from a psychological phenomenon called Ambiguity Aversion \citep{fox1995ambiguity, maccheroni2006ambiguity} where people tend to answer more conservatively when encountering risks and uncertainty, ensuring the lower bound of the gain. Formally, denoting $s$ as a non-empty subset of the label space $\mathcal{Y}$, the prediction for $x$ given by LLM is defined as:
% \begin{equation}
% \label{eq:crg}
%     \underset{s \subseteq \mathcal{Y}, s \neq \emptyset}{\argmax} ~ P\left(V'\left(s\right) | \{T(x_i^{d}, \tilde{y}_i^{d})\}_{i=1}^m, T'\left(x\right)\right)
% \end{equation}
% where $T'(\cdot)$ is a natural language template querying LLM a candidate label set for each sample and $V'(\cdot)$ maps a set of labels to a concatenation of their pre-defined class tokens. 

\begin{figure*}[!t]
    \centering
    \includegraphics[width=0.99\linewidth]{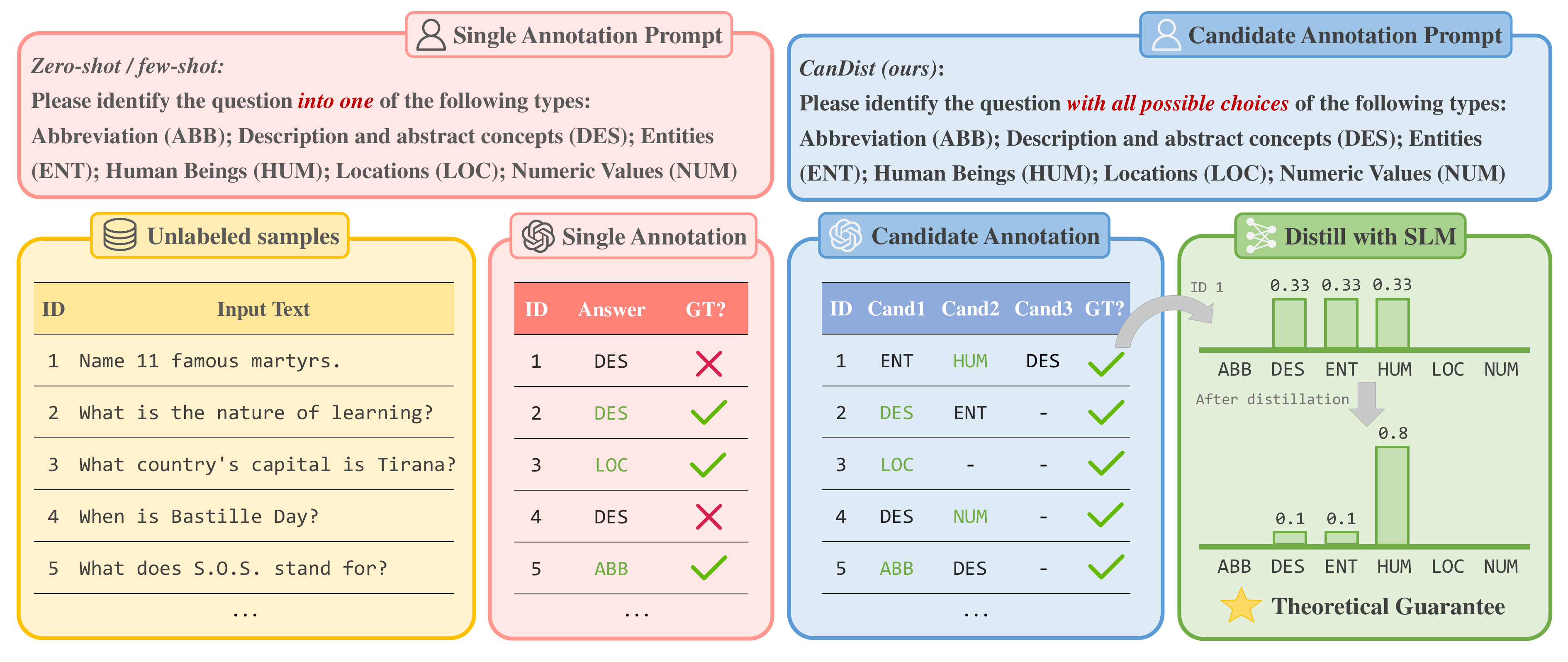}
    \caption{The overall framework of CanDist, which first prompts the LLM to provide candidate annotations, and then distills an SLM to identify the correct labels. Examples on the TREC dataset annotated by GPT-3.5 demonstrate that though the LLM fails to provide a correct answer with a single label, answering with candidate labels successfully includes the correct one. We also provide theoretical guarantees for our proposed CanDist framework.}
    \label{fig:framework}
\end{figure*}

However, LLM's knowledge of downstream tasks remains limited \citep{DBLP:conf/acl/LiCRCZNW24}, making them uncertain about some samples during data annotation. In this case, prompting with single annotations may force the LLM to behave over-confidently and generate completely incorrect answers, which not only wastes computational resources but also harms downstream processes. To tackle this problem, we propose to prompt LLM with \textbf{Candidate Annotations} (CA), namely, a set of multiple possible labels $s \subseteq \mathcal{Y}, s \neq \emptyset$. Our motivation stems from a human psychological phenomenon known as Ambiguity Aversion \citep{fox1995ambiguity, maccheroni2006ambiguity}, where people tend to behave conservatively when facing uncertainty, which helps mitigate severe risks and ensures the lower bound of the gains. Prompting candidate annotations can inject ambiguity aversion into LLMs, which increases the likelihood of including correct labels in LLM's output, see examples in Figure \ref{fig:framework}. 

Specifically, we investigate two strategies for querying candidates: 1) $\text{CA}_\text{add}$ prompts the LLM to generate one answer first and then provide additional answers if it is not sure; 2) $\text{CA}_\text{all}$ prompts the LLM to generate all possible answers.
Table \ref{tab:prompts} shows the key prompts of different prompting strategies on the TREC dataset and the full prompts can be found in Appendix \ref{subsec:full_prompts}.

\paragraph{CA Exhibits Better Statistical Properties.}
In this paragraph, we directly assess the value of candidate annotations.
Regarding the annotation process as label space pruning, we employ the metrics introduced in \citep{DBLP:conf/iclr/0001WY024}: 1) $1-\alpha$-error, where $\alpha=\frac{1}{n}\sum_{i=1}^n \mathbb{I}[y_i \notin s_i]$, measuring how the candidates include the correct labels; 2) $\beta$-coverage, where $\beta=\frac{1}{n}\sum_{i=1}^n \frac{C-|s_i|}{C-1}$, measuring how the answers shrink the original search space; 3) F1-score, which comprehensively considers both metrics, namely, $\text{F1}=\frac{2(1-\alpha)\beta}{1-\alpha+\beta}$.

Figure \ref{fig:gpt_assessment} demonstrates the assessment results of $1-\alpha$-error and F1-score on three text classification tasks annotated by GPT-3.5, where both $\text{CA}_\text{add}$ and $\text{CA}_\text{all}$ improves the two metrics compared to SA. Notably, by prompting all possible labels, $\text{CA}_\text{all}$ outperforms SA by margins of \textbf{18.01\%}, \textbf{26.71\%}, \textbf{14.06\%} of $1-\alpha$-error on the three datasets, indicating the strong ability to include gold labels of prompting candidate annotations. The higher F1-scores further illustrate that while containing more correct labels, CA also effectively shrinks the search space, indicating its great value. The full assessment results are in Appendix \ref{subsec:full_assessment}.
% For the more comprehensive metric F1-score, $\text{CanDist}$ consistently outperforms few-shot inference with an improvement ranging from 1.88\% to 8.88\%. The above statistical results reveal the great value embodied in candidate annotation. 

% \begin{figure}[!t]
%     \centering
%     \begin{subfigure}{0.48\textwidth}
%         \centering
%         \includegraphics[width=0.85\textwidth]{img/gpt-fs-training-acc-3.png}
%         % \caption{Precision.}
%         % \label{fig:llama_testing_acc}
%     \end{subfigure}
%     \begin{subfigure}{0.48\textwidth}
%         \centering
%         \includegraphics[width=0.85\textwidth]{img/gpt-fs-training-f1-3.png}
%         % \caption{F1-score.}
%         % \label{fig:llama_testing_f1}
%     \end{subfigure}
%     \caption{Comparison of Precision ($1-\alpha$-error) and F1-score between different prompting strategies on 3 training datasets using GPT-3.5. Higher metric values indicate better results.}
%     \label{fig:gpt_assessment}
% \end{figure}

\subsection{Distill Candidate Annotations by SLM}\label{subsec:learning}

% Previous approaches have shown that distilling SLMs from LLM-generated data can boost LLMs' performance on specific tasks \citep{DBLP:conf/acl/MagisterMAMS23, DBLP:conf/icml/FuPOSK23, DBLP:conf/emnlp/0001YMWRCYR23, DBLP:conf/emnlp/Xiao0ZWLCW23}. 
% FreeAL \citep{DBLP:conf/emnlp/Xiao0ZWLCW23}, one of the pioneer works, distills the LLM-generated single annotation to SLMs by a robust training objective, regarding each single annotation as a weak label. Unlike single annotation, distilling from candidate annotation requires customized training objectives.
Though candidate annotations demonstrate great potential, they cannot be directly applied to downstream tasks where specific labels are required.
% The most intuitive solution would be to let the LLM themselves or humans select from the candidate annotations. However, since LLMs are yet challenging to self-correct \citep{DBLP:conf/iclr/0009CMZYSZ24} and the cost of human labeling prohibitively high, we believe these are not the optimal solutions.
To address this, we propose a teacher-student framework \textbf{CanDist} that trains an SLM student $\mathcal{S}$ on the teacher LLM's candidate annotations, allowing the SLM to provide unique annotations.
This is inspired by knowledge distillation \citep{DBLP:journals/corr/HintonVD15}, where the student model is distilled from the teacher model's output distribution and can better generalize to downstream tasks \citep{DBLP:conf/icml/PhuongL19,DBLP:conf/iclr/JeongC25}. The overall framework of CanDist is shown in Figure \ref{fig:framework}.

Nevertheless, with multiple false positive labels, training the SLM on the uniform distribution of candidate labels is suboptimal. Therefore, we propose a Distribution Refinery (DR) strategy, which dynamically adjusts the target distribution based on the SLM's prediction. This is motivated by the memorization effect of deep neural networks (DNNs) \citep{DBLP:conf/iclr/ZhangBHRV17}, where the SLM can first remember easy patterns, making a proportion of true labels emerge from those false positive ones. Formally, the refined distribution $\bm{q}_i$ for sample $\bm{x}_i$ at each training iteration $t$ is computed as the re-normalized prediction among candidate labels:
\begin{equation}
\label{eq:rn}
  \begin{aligned}
    % \mathcal{L}_{rw} &= \frac{1}{n}\sum_{i=1}^n l_\text{ce}(\bm{f}^t(\bm{x}_i), \bm{q}_i^t), \\
    q_{ij}^t =& \begin{cases}
        \mathbb{I}(j \in s_i) \cdot \frac{1}{|s_i|}, & t = 0 \\
        \mathbb{I}(j \in s_i) \cdot p_{ij}^{t-1} / \sum_{k \in s_i} p_{ik}^{t-1}, & t > 0 \\
    \end{cases}
  \end{aligned}
\end{equation}
where $\bm{p}^t_i$ denotes the SLM's softmax output of sample $\bm{x}_i$ at iteration $t$. $\bm{q}_i$ is the distribution vector which is initialized from a uniform distribution.

\paragraph{Filter Out-of-Candidate Samples.}
Although candidate annotations are more likely to include the correct labels, there are still a few samples whose true label lies outside the candidate set, which can disrupt SLM distillation.
To this end, we filter out these samples by judging whether the SLM's max prediction lies beyond the candidate set:
\begin{equation}
    \label{eq:out}
    \begin{aligned}
        & \mathcal{D}_\text{out} = \{ \bm{x}_i | \mathop{\argmax}_{c \in \mathcal{Y}} p_{ic} \notin s_i \} \\
        % & \mathcal{D}_\text{in} = \mathcal{D}-\mathcal{D}_\text{out}
    \end{aligned}
\end{equation}

\begin{algorithm}[!t]
\renewcommand{\algorithmicrequire}{\textbf{Input:}}
\renewcommand{\algorithmicensure}{\textbf{Output:}}
\caption{Pseudo-code of CanDist}
\label{alg}
\textbf{Input:} Unlabeled dataset $\mathcal{D}$, teacher LLM $\mathcal{T}$, and student SLM $\mathcal{S}$
    \begin{algorithmic}[1] %[1] enables line numbers	
    \STATE Generate candidate annotations $s$ using $\mathcal{T}$ by prompting strategy $\text{CA}_\text{add}$ or $\text{CA}_\text{all}$ \\
    \FOR {$epoch=1,2,\ldots,$}
        \STATE Filter out-of-candidate samples by Eq.(\ref{eq:out}) \\
        \STATE Select class-wise reliable samples by Eq.(\ref{eq:cwsl}) \\
        \STATE Select high confidence samples by Eq.(\ref{eq:hc}) \\
        \FOR {$batch=1,2,\ldots,$}
            \STATE Compute pseudo-labels by Eq.(\ref{eq:rn}) and (\ref{eq:final}) \\
            \STATE Calculate training loss $\mathcal{L}_{dr}$ by Eq.(\ref{eq:final}) \\
            \STATE Train $\mathcal{S}$ by optimizing $\mathcal{L}_{dr}$ \\
        \ENDFOR
    \ENDFOR
    \end{algorithmic}
\textbf{Output:} {Student SLM $\mathcal{S}$ for annotation}
\end{algorithm}

\paragraph{Distribution Sharpening for Reliable Samples.}
We further propose to select reliable samples in $\mathcal{D}_\text{in} = \mathcal{D}-\mathcal{D}_\text{out}$ and sharpen their target distributions to guide the distillation process.
To assess the reliability, we again leverage the memorization effect of DNNs where clean samples always pose small losses \citep{DBLP:conf/nips/HanYYNXHTS18}. Specifically, we select small loss samples in a class-wise manner to ensure balanced training progress across all classes. Formally, the reliable set is calculated as:
\begin{equation}
\label{eq:cwsl}
\begin{aligned}
    & \mathcal{D}_\text{sl} = \mathop{\cup}_{c\in\mathcal{Y}} \mathcal{D}^c_\text{sl},~\text{where} \\
    & \mathcal{D}^c_\text{sl} = \{ \bm{x}_i | l_i\in\mathcal{L}^c_\delta, l_i=l_\text{ce}(\bm{p}_i, \bm{q}_i) \} \\
    % \mathcal{D}^c = \{\bm{x}_i | c=\underset{{j \in s_i}}{\argmax}~ p_{ij}\} \cap \mathcal{D}_\text{in} \\
\end{aligned}
\end{equation}
and $l_\text{ce}$ denotes the cross-entropy loss, and $\mathcal{L}^c_\delta$ denotes the top-$\delta$ percent smallest losses of samples whose max prediction is class $c$. For samples in $\mathcal{D}_\text{sl}$, we use a pre-defined temperature $\gamma$ to sharpen their re-normalized distribution. 

Besides, we regard those samples in $\mathcal{D}_\text{out}$ that gradually pose high confidence as reliable samples:
\begin{equation}
\label{eq:hc}
    \mathcal{D}_\text{hc}=\{\bm{x}_i | \max_{c \in \mathcal{Y}} p_{ic} > \tau \} \subset \mathcal{D}_\text{out}
\end{equation}
where we use their predicted class as the training target. $\tau$ is a pre-defined high threshold. 

% $\gamma=\gamma_1$ and the rest with $\gamma=\gamma_2>\gamma_1$. The rationality behind such class-wise assignments is that LLM suffers from biases during selection from multiple choices \citep{DBLP:journals/corr/abs-2308-11483, DBLP:conf/iclr/Zheng0M0H24} leading to long-tailed data annotation, and class-wise assignment enables a balanced learning process for different classes \citep{DBLP:conf/nips/WangXLM00Z22}.

\paragraph{Overall Distillation Object.} The overall training objective of Distribution Refinery is formalized as:
\begin{equation}
\label{eq:final}
  \begin{aligned}
    & \mathcal{L}_\text{dr} = \frac{1}{n}\sum_{i=1}^n l_\text{ce}(\bm{p}_i, \hat{\bm{q}}_i), ~\text{where}\\
    \hat{q}_{ij} &= \begin{cases}
        q_{ij}^{1/\gamma} / \sum_{c \in \mathcal{Y}} q_{ic}^{1/\gamma}, & \bm{x}_i \in \mathcal{D}_\text{sl} \\
        q_{ij}, & \bm{x}_i \in \mathcal{D}_\text{in}-\mathcal{D}_\text{sl} \\
        \mathbb{I}(j = \underset{c\in\mathcal{Y}}{\argmax}~ p_{ij}), & \bm{x}_i \in \mathcal{D}_\text{hc} \\
    \end{cases}
  \end{aligned}
\end{equation}
Algorithm \ref{alg} shows the pseudo-code of CanDist.

\section{Theoretical Analysis}

In this section, we further theoretically explain why prompting and then distilling candidate annotations leads to better results. 
Since there is still a lack of theoretical understanding of LLMs, we simplify this problem by treating the LLM as a traditional teacher model, focusing on whether the SLM can distill better results from candidate labels.
While most existing knowledge distillation theories illustrate the advantages of distilling from the teacher's output distribution \citep{DBLP:conf/icml/PhuongL19, DBLP:conf/icml/DasS23}, we analyze distilling from the teacher's candidate annotations (top-$k$ outputs), wherein the student SLM distilled from teacher LLM's candidate annotations demonstrate more noise-tolerant than the teacher LLM, as well as the SLM distilled from LLM's single annotations. 

\begin{table*}[!t]
    \caption{Comparisons of Accuracies (\%) on the training and testing sets of different tasks. $\text{CanDist}_\text{add}$ and $\text{CanDist}_\text{all}$ apply $\text{CA}_\text{add}$ and $\text{CA}_\text{all}$ to prompt candidates respectively. The best results are bold and the second best is underlined.}
    \label{tab:main}
    \centering
    \renewcommand\arraystretch{1.20}
    \scalebox{0.8}{
    \begin{tabular}{l|p{1.01cm}<{\centering}p{1.01cm}<{\centering}p{1.01cm}<{\centering}p{1.01cm}<{\centering}p{1.01cm}<{\centering}p{1.01cm}<{\centering}|p{1.01cm}<{\centering}p{1.01cm}<{\centering}p{1.01cm}<{\centering}p{1.01cm}<{\centering}p{1.01cm}<{\centering}p{1.01cm}<{\centering}}
    \toprule
        \multirow{2}{*}{Method} & \multicolumn{6}{c|}{Training Set} & \multicolumn{6}{c}{Testing Set} \\
    \cmidrule(r){2-13}
        ~ & \textbf{TREC} & \textbf{MA} & \textbf{DBP} &  \textbf{AGN} & \textbf{RCT} & \textbf{BANK} & \textbf{TREC} & \textbf{MA} & \textbf{DBP} &  \textbf{AGN} & \textbf{RCT} & \textbf{BANK} \\ 
    \midrule
        Zero-shot & 62.84 & 62.03 & 93.33 & 87.72 & 61.41 & 65.19 & 72.20 & 63.12 & 93.94 & 87.24 & 61.83 & 68.41 \\ 
        Few-shot & 71.07 & 62.28 & 95.41 & 88.73 & 65.18 & 66.08 & 77.20 & 63.40 & 95.40 & 88.05 & 65.85 & 68.86 \\ 
        CoT & 71.88 & 60.05 & 91.85 & 83.23 & 60.06 & 57.54 & 80.60 & 61.15 & 92.44 & 83.05 & 60.43 & 60.97 \\ 
        SC & 71.06 & 62.29 & 95.60 & 88.80 & 65.50 & 66.08 & 76.00 & 63.26 & 95.42 & 87.96 & 65.85 & 68.99 \\ 
        AnnoLLM & 73.73 & 59.71 & 95.62 & 85.52 & 68.13 & 67.04 & 79.60 & 59.56 & 95.34 & 85.39 & 68.53 & 70.29 \\ 
        SuperICL & 76.05 & 62.81 & 97.55 & 89.16 & 66.80 & 69.91 & 81.60 & 63.75 & 97.63 & \underline{88.79} & 67.82 & 73.25 \\ 
        Distillation & 76.04 & 62.45 & 97.52 & 89.13 & 66.86 & 69.83 & 81.00 & 63.54 & 97.61 & 88.29 & 67.66 & 72.40 \\ 
        FreeAL & 78.24 & 62.89 & 97.76 & \underline{89.58} & 67.57 & 71.38 & 82.33 & 64.13 & 97.92 & 88.64 & 68.32 & 74.58 \\ 
        \rowcolor{gray!20}
        $\text{CanDist}_\text{add}$ & \textbf{80.87} & \underline{63.31} & \textbf{98.67} & \textbf{89.91} & \underline{68.69} & \underline{72.92} & \underline{83.13} & \textbf{64.23} & \textbf{98.72} & \textbf{89.46} & \underline{69.77} & \textbf{76.27} \\   
        \rowcolor{gray!20}
        $\text{CanDist}_\text{all}$ & \underline{79.73} & \textbf{63.76} & \underline{98.54} & 89.29 & \textbf{68.90} & \textbf{72.94} & \textbf{87.80} & \underline{64.20} & \underline{98.65} & 88.78 & \textbf{70.57} & \underline{75.97} \\ 
    \midrule
        SFT & - & - & - & - & - & - & \textcolor{black!60}{97.80} & \textcolor{black!60}{64.54} & \textcolor{black!60}{98.78} & \textcolor{black!60}{92.29} & \textcolor{black!60}{84.52} & \textcolor{black!60}{93.31} \\ 
    \bottomrule
    \end{tabular}}
\end{table*}

\begin{theorem}
\label{thm}
Considering the scenario that both the teacher LLM and student SLM are composed of a feature extractor $\bm{g}(\cdot):\mathcal{X}\mapsto \mathbb{R}^d$ (with different scales) and a classifier $\bm{W}\in\mathbb{R}^{d\times C}$. The teacher LLM is pre-trained on an inaccurate dataset $\tilde{\mathcal{D}}=\{\bm{x}_i, \tilde{y}_i \}_{i=1}^m$ with noise rates $\{\bm{R}_{c, c'}\}_{c=1, c'=1}^{C,C}$\footnote{Due to LLMs' strong general capabilities, we assume that, for a specific task, LLMs can consistently output a label distribution $P'$ that is relatively close to the true distribution $P$. Under this assumption, LLMs appear to act like a teacher pre-trained on a dataset with distribution $P'$.}, where $m$ denotes the number of samples in the dataset and $\bm{R}_{c, c'}$ indicates the probability of label $c$ being flipped to $c'$. After pre-training, the student SLM is then trained based on the teacher LLM's single (top-1) or candidate (top-2) annotations on $\tilde{\mathcal{D}}$. Suppose the models are trained by $l_2$-regularized cross-entropy loss with regularization parameter $\lambda$, and the feature extractors are fixed. Besides, we consider that the feature similarity between different samples from the same class and different classes are $a$ and $b$ respectively, with $1>a>b>0$.

Then, with $m\to \infty$, the condition of achieving 100\% accuracy (correctly predicting all training data) for the teacher LLM, as well as the student SLM distilled from LLM's top-1 prediction is:
\begin{equation}
\begin{gathered}
    \bm{R}_{c, c'} + \sum_{i\neq c}\bm{R}_{c, i} < 1 - \frac{\theta}{\phi-\theta},~\forall c, c'\neq c \\ 
    \text{where} ~ \theta = 1-\frac{Cm\lambda}{Cm\lambda+1-a}, \\
    \phi = 1 - \frac{Cm\lambda}{Cm\lambda + \frac{m}{C}(a-b)+1-a} \\
\end{gathered} 
\end{equation}
and the condition of that for the student SLM distilled from LLM's top-2 prediction is:
\begin{equation}
    \bm{R}_{c, c'} + \sum_{i\neq c}\bm{R}_{c, i} < 1 ,~\forall c,c'\neq c
\end{equation}
\end{theorem}
The proof is provided in Appendix \ref{sec:proof}. The theorem illustrates that the SLM distilling top-2 predictions from the teacher LLM achieves 100\% accuracy with \textbf{a more tolerant condition on label noise} than using the top-1 prediction, which theoretically demonstrates the great potential of the paradigm that first generates candidates by the teacher LLM and then distilling them using a student SLM.

\section{Experiments}

In this section, we report our empirical results to show the superiority of CanDist. We refer the readers to the Appendix for more details and results.

\subsection{Setup}
\paragraph{Datasets.} We conduct experiments on the following six text classification datasets, namely, \textbf{TREC} \citep{DBLP:conf/coling/LiR02} for topic classification, Medical Abstract (\textbf{MA}) \citep{DBLP:conf/nlpir/Schopf0M22} for medical diagnosis classification,  DBpedia (\textbf{DBP}) for ontology classification \citep{DBLP:conf/nips/ZhangZL15}, AGNews (\textbf{AGN}) \citep{DBLP:conf/www/Gulli05} for news topic classification, \textbf{RCT} \citep{DBLP:conf/ijcnlp/DernoncourtL17} for content type classification in medical abstracts, and Banking (\textbf{BANK}) \citep{DBLP:journals/corr/abs-2003-04807} for intent classification in banking dialogues. 

\paragraph{Baselines.} We adopt the following LLM-based or SLM-based baselines: \textbf{Zero-shot} and \textbf{Few-shot} \citep{DBLP:conf/acl-deelio/LiuSZDCC22} directly prompt for single annotations without/with few-shot examples; \textbf{CoT} \citep{DBLP:conf/nips/KojimaGRMI22} employs chain-of-thought prompting by adding "Let's think step by step" before each answer; Self-Consistency (\textbf{SC}) \citep{DBLP:conf/iclr/0002WSLCNCZ23} samples diverse reasoning paths and generates the answer by majority voting; \textbf{AnnoLLM} \citep{DBLP:conf/naacl/HeLGJZLJYDC24} provides explanations for few-shot examples to boost performance; \textbf{SuperICL} \citep{DBLP:conf/acl/XuXWLZM24} first trains an SLM using labeled data and uses its output and confidence as references during LLM annotation; \textbf{Distillation} distill an SLM from LLM's single annotation and use the SLM to provide the final annotation; \textbf{FreeAL} \citep{DBLP:conf/emnlp/Xiao0ZWLCW23} introduces a robust training mechanism to improve generalization when distilling the SLM from single annotations, where we apply 1 round of annotation-distillation for a fair comparison. Note that few-shot examples are applied to CanDist and all baselines except Zero-shot and CoT. Besides, for SuperICL, LLM's single annotations are leveraged to train the plug-in SLM.

\paragraph{Performance Evaluation.}
We evaluate the annotation accuracy on both the training and testing set. For SLM-based methods (Distillation, FreeAL, and our method), the unlabeled training set is first annotated by the LLM, and then the SLM is trained on this training set to provide annotations.
We also report the testing results of supervised fine-tuning (\textbf{SFT}) where the SLM is trained on the human-labeled training dataset.
For all experiments, we run three times and report the averaged results.

\begin{table*}[!t]
    \caption{Comparison with selecting answers from candidates using LLM on the training sets. Results of single annotations (Few-shot) are also listed for the sake of comparison.}
    \label{tab:ab_identify}
    \centering
    \renewcommand\arraystretch{1.15}
    \scalebox{0.79}{
    \begin{tabular}{l|p{2.01cm}p{2.05cm}p{2.01cm}p{2.01cm}p{2.01cm}p{2.01cm}|p{2.01cm}}
    \toprule
        \textbf{Ablation} & \textbf{TREC} & \textbf{MA} & \textbf{DBP} & \textbf{AGN}& \textbf{RCT} & \textbf{BANK} & \textbf{Avg.} \\ 
    \midrule
        \textbf{$\text{CanDist}_\text{add}$} & 80.87 & 63.31 & 98.67 & 89.91 & 68.69 & 73.50 & 79.16 \\ 
        % w/o Dist. Refinery & 79.20 \green{(-1.67)} & 61.97 \green{(-1.33)} & 72.24 \green{(-1.26)} & 88.74 \green{(-1.17)} & 97.18 \green{(-1.49)} \\ 
        with LLM Select & 72.87 \green{(-8.00)} & 63.42 \red{(+0.11)} & 96.38 \green{(-2.29)} & 88.33 \green{(-1.58)} & 63.17 \green{(-5.52)} & 68.33 \green{(-5.16)} & 75.42 \green{(-3.74)} \\
    \midrule
        \textbf{$\text{CanDist}_\text{all}$} & 79.73 & 63.76 & 98.54 & 89.29 & 68.90 & 72.94 & 78.86 \\ 
        % w/o Dist. Refinery & 74.96 \green{(-4.77)} & 61.34 \green{(-2.42)} & 94.67 \green{(-3.87)} & 84.37 \green{(-4.92)} & 69.14 \green{(-3.80)} & 76.90 \green{(-3.95)} \\ 
        with LLM Select & 70.95 \green{(-8.78)} & 63.18 \green{(-0.58)} & 96.30 \green{(-2.24)} & 88.23 \green{(-1.06)} & 63.67 \green{(-5.23)} & 67.42 \green{(-5.52)} & 74.96 \green{(-3.90)} \\
    \midrule
        Few-shot & 71.07 & 62.28 & 95.41 & 88.73 & 65.18 & 66.08 & 74.79 \\
    \bottomrule
    \end{tabular}}
\end{table*}

\paragraph{Implementation Details.} We exploit GPT-3.5 as the LLM annotator (see results of more advanced LLMs in Appendix \ref{subsec:4o}) and RoBERTa-Base \citep{DBLP:journals/corr/abs-1907-11692} as the SLM for all tasks except MA, where BioMed-RoBERTa-Base \citep{DBLP:conf/acl/GururanganMSLBD20} is used to boost performance for the medical task.
We set the number of few-shot examples as 10 for all tasks except 5 for MA due to limited context length. Since we cannot access labeled samples, the few-shot examples are LLM-generated \citep{DBLP:conf/emnlp/Xiao0ZWLCW23}. 
For sampling-based baseline SC, we sample the decoding path 5 times with a temperature of 0.5. For other LLM generation processes, the temperature is set to a lower value of 0.3. More details of training SLM are in Appendix \ref{subsec:details}.

\subsection{Main Results}
The comparison results on the training and testing sets are shown in Table \ref{tab:main} where the best results are shown in bold and the second best is underlined. Overall, CanDist outperforms all baselines on all tasks. For example, on the testing set of TREC, CanDist improves the best baseline by a large margin of \textbf{5.47\%}. Also, in the tasks of MA and DBpedia, CanDist achieves competitive testing performance \textbf{on par with supervised fine-tuning}. The superior results against all baselines imply the effectiveness of our proposed CanDist framework.

Specifically, CanDist largely improves Zero-shot and Few-shot, where $\text{CanDist}_\text{add}$ and $\text{CanDist}_\text{all}$ outperform Few-shot by averaged improvements of 5.48\% and 6.10\% on the testing set, and 7.03\% and 6.63\% on the training set. Though effective in reasoning tasks, CoT prompting performs poorly in most annotation tasks and self-consistency achieves similar results with Few-shot. AnnoLLM improves Few-shot in several tasks by providing explanations on input examples. However, these LLM-based methods underperform SLM-based methods, where SLM can distill the high-quality task-related knowledge from the LLM's annotation. Regarding the knowledge of SLM as a reference, SuperICL slightly improves the performance of Distillation. FreeAL further improves Distillation through a robust training objective that tackles label noise.
For CanDist, we declare that there is a trade-off between the number of candidates and the accuracy since more candidates are more challenging to identify while fewer candidates contain fewer correct labels. Though $\text{CA}_\text{all}$ generally retrieves more labels than $\text{CA}_\text{add}$, we suppose that the performance of different prompting strategies depends on tasks, and both strategies achieve state-of-the-art results.

\begin{table}[!t]
    \caption{Comparison with other candidate generation strategies on TREC, where $1-\alpha$-error, average number of labels (\#Labels), and testing accuracy are reported}
    \label{tab:ab_generate}
    \centering
    \renewcommand\arraystretch{1.05}
    \scalebox{0.9}{
    \begin{tabular}{l|ccc}
    \toprule
        \textbf{Strategy} & $1-\alpha$ & \textbf{\#Labels} & \textbf{Accuracy} \\
    \midrule
        5 sampled paths & 77.59 & 1.17  & 81.40 \\ 
        10 sampled paths & 79.92 & 1.25  & 81.73 \\ 
        20 sampled paths & 82.36 & 1.32  & 82.27 \\ 
        40 sampled paths & 84.30 & 1.39  & 82.33 \\ 
    \midrule
        5 example orders & 79.15 & 1.21 & 81.27 \\ 
        5 prompt formats & 83.82 & 1.30 & 82.67 \\
    \midrule
        $\text{CanDist}_\text{add}$ & 74.65 & 1.07  & \underline{83.13} \\
        $\text{CanDist}_\text{all}$ & 89.09 & 1.70  & \textbf{87.80} \\ 
    \bottomrule
    \end{tabular}}
\end{table}

\subsection{Analysis}

\paragraph{Comparison with Other Candidate Generation Strategies.} To show the superiority of generating candidates by prompting, we compare the following two candidate generation strategies: 1) \textit{sampling-based} strategy \citep{DBLP:conf/iclr/0002WSLCNCZ23} samples $K=5, 10, 20, 40$ paths and gathers them into a candidate set; 2) \textit{ensemble-based} strategy gathers the answers from diverse prompting results, where we consider prompting with 5 few-shot example orders \cite{DBLP:conf/icml/ZhaoWFK021} and 5 prompting formats \citep{DBLP:conf/acl/GaoFC20}. To evaluate the generated candidates, we report their $1-\alpha$-error, average number of labels, and the testing accuracy of SLM trained by our proposed 
Distribution Refinery objective.

Table \ref{tab:ab_generate} demonstrates that by retrieving more candidate labels, $\text{CanDist}_\text{all}$ enjoys much higher $1-\alpha$-error than other methods and achieves the highest testing accuracy.
Moreover, $\text{CanDist}_\text{add}$ also outperforms the sampling and ensemble-based methods even if it retrieves fewer candidates, indicating that directly prompting candidates results in more valuable annotation. 
For the sampling-based method, though incorporating more sampled paths offers a higher $1-\alpha$-error, the increment in testing accuracy remains limited. Besides, sampling and ensemble-based strategies suffer from more costs in querying LLMs while promoting candidates only need to prompt and sample once.

\begin{table}[!t]
    \caption{Key prompt for selecting answers from candidate annotations on the TREC dataset.}
    \label{tab:prompt_select}
    \centering
    \small
    \renewcommand\arraystretch{1.0}
    \scalebox{1.0}{
    \begin{tabular}{p{7cm}}
    \toprule
        \textbf{Prompt of selecting the answer from candidates} \\ 
    \midrule
        \parbox{7cm}{\textsf{Given a question: $\dots$ What does this question ask about? It is known that the answer belongs to one of the following classes: .... Please select the correct answer from them.}} \\ 
    \bottomrule
    \end{tabular}}
\end{table}

\paragraph{Comparison with Selecting Answers from Candidates using LLM.}
To validate the effectiveness of our proposed teacher-student framework for identifying the correct label from candidate labels, we compare CanDist with its variant, \textit{CanDist with LLM Select}, which directly queries LLM to select the correct label from the given candidate annotations. The key prompt for selecting the answer from candidates is shown in Table \ref{tab:prompt_select}. As shown in Table \ref{tab:ab_identify}, LLM selection suffers from performance drops compared with CanDist on most tasks, which demonstrates the superiority of our proposed teacher-student framework. Moreover, we found that CanDist with LLM Selection slightly outperforms single annotations (Few-shot), indicating that the paradigm of prompting candidates and then selecting from them is better than direct prompting for a single label.

\begin{table}[!t]
    \caption{Ablation study on Distribution Refinery mechanism on the testing set of TREC and Banking.}
    \label{tab:ab_apl}
    \centering
    \renewcommand\arraystretch{1.1}
    \scalebox{0.83}{
    \begin{tabular}{ccccc|p{1.21cm}<{\centering}p{1.21cm}<{\centering}}
    \toprule
        Ren. & Out. & Sha. & Cla. & Hig. & \textbf{TREC} & \textbf{BANK} \\ 
    \midrule
        ~ & ~ & ~ & ~ & ~ & 82.47 & 71.40 \\ 
        \checkmark & ~ & ~ & ~ & ~ & 85.47 & 74.88 \\ 
        \checkmark & \checkmark & ~ & ~ & ~ & 86.60 & 75.13 \\ 
        \checkmark & \checkmark & \checkmark & ~ & ~ & 87.07 & 74.99 \\ 
        \checkmark & \checkmark & \checkmark & \checkmark & ~ & 87.40 & 75.70 \\ 
        \checkmark & \checkmark & \checkmark & \checkmark & \checkmark & \textbf{87.80} & \textbf{75.97} \\   
    \bottomrule
    \end{tabular}}
\end{table}

\paragraph{Ablation Study on Distribution Refinery.} To demonstrate the effectiveness of different components in DR, we run $\text{CanDist}_\text{all}$ with varying combinations of the components. We denote the components in DR as 1) \textit{Ren.} for the re-normalization function in Eq.(\ref{eq:rn}); 2) \textit{Out.} for filtering out-of-candidate samples; 3) \textit{Sha.} for whether employing distribution sharpening for reliable samples; 4) \textit{Cla.} for whether select small loss samples in a class-wise manner; 5) \textit{Hig.} for whether using high confidence samples as reliable samples. As shown in Table \ref{tab:ab_apl}, distilling from re-normalized distribution improves the vanilla version (trained on cross-entropy loss) by a large margin, i.e., 3.00\% for TREC and 3.70\% for Banking. DR also helps by filtering out-of-candidate samples and sharpening the target distribution, where class-wise selection is essential for employing distribution sharpening, which balances the training progress across all classes. High-confidence label assignment further improves the performance by maximizing the utility of the out-of-candidate samples.

\begin{figure}[!t]
    \centering
    \includegraphics[width=0.99\linewidth]{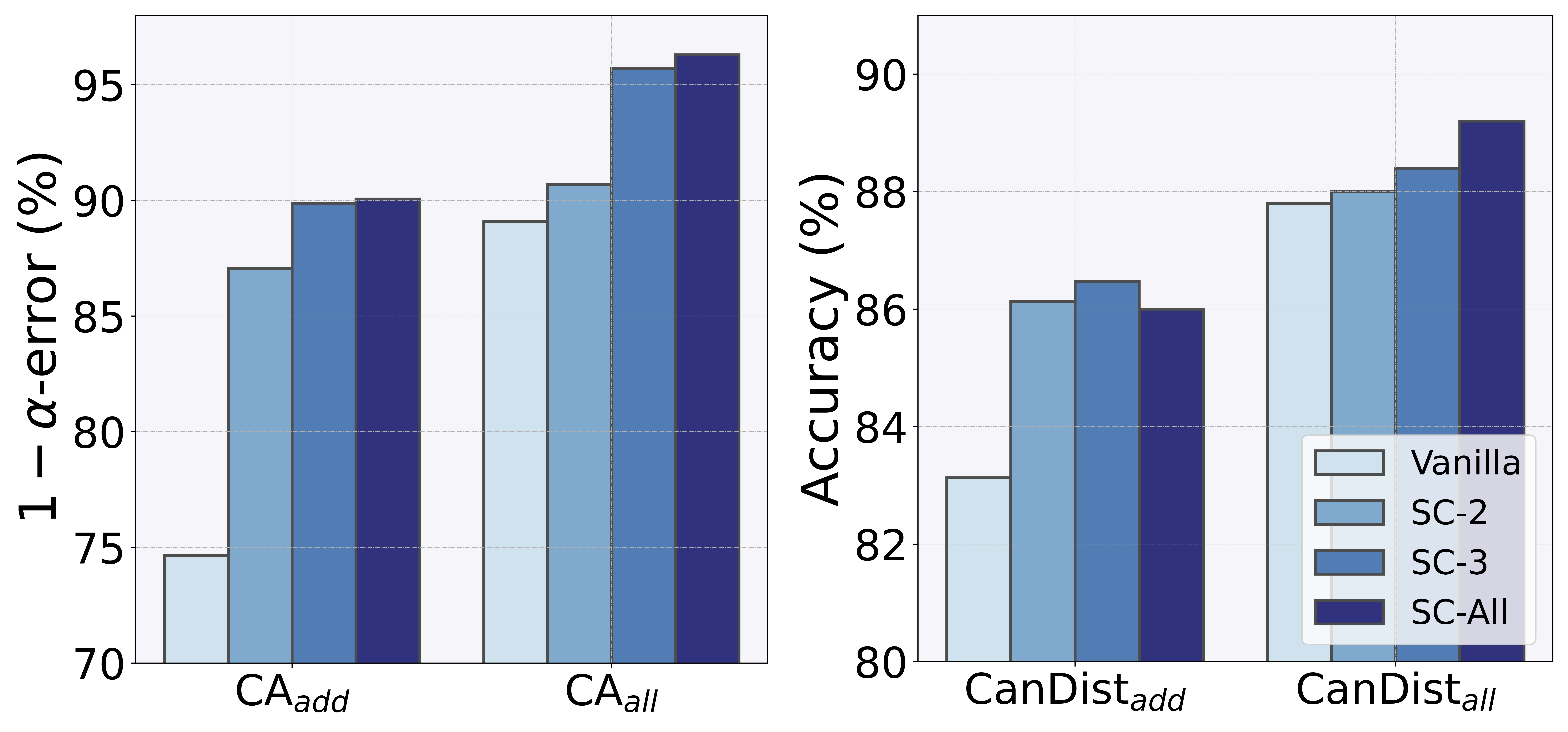}
    \caption{Comparison of $1-\alpha$ on TREC's training set (left) and accuracy on the testing set (right) between different collaboration strategies with self-consistency.}
    \label{fig:ablation_scmv}
\end{figure}

\paragraph{Synergism with Self-Consistency.}\label{para:scmv} We further show that our vanilla method can work collaboratively with Self-Consistency (SC). Specifically, we first prompt LLMs with candidate labels and sample $K=40$ answers $\{s_j\}_{j=1}^K$, and then calculate the frequency for each class $c$ by $\sum_{j=1}^K \mathbb{I}(c \in s_j)$ to filter the top-$k$ frequent labels as candidate annotations. We name this strategy as \textit{SC-$k$} and we also define \textit{SC-All} as using all the appeared labels as candidate labels. As shown in Figure \ref{fig:ablation_scmv}, the comparison on $1-\alpha$-error illustrates that collaborating with SC further increases the diversity of candidate labels which includes more correct labels, and this also yields a higher accuracy for the final annotation, as shown on the right. Further discussion on SC-1 can be found in Appendix \ref{subsec:mv}.

\section{Conclusion}
In this work, we study LLM-driven data annotation by proposing a novel teacher-student framework, CanDist, which first prompts the teacher LLM to generate candidate labels and then distill a student SLM to identify the true labels.
We illustrate that candidate annotations exhibit better statistical properties and theoretically justify that distilling from LLM's candidate annotations is more noise-tolerant. Empirically, we show that CanDist outperforms various LLM and SLM-based methods. We hope our work will inspire future research to exploit candidate annotations with weak annotators.

\section*{Limitations}
Despite the effectiveness of our proposed CanDist framework for data annotation, there is still much potential for further improvement. On the one hand, as the Distribution Refinery mechanism is specifically designed for classification, the application of CanDist is currently limited to text classification tasks, and we aim to explore its potential in text generation tasks in our future works. On the other hand, the derivation of our proposed theory is based on the assumption that the LLM is a traditional encoder model, which is not the case for the prevailing decoder-only LLMs. Besides, there is still a lack of theoretical understanding of LLMs in the community and we hope that this field will further develop in the near future.

\section*{Ethical Considerations}
While the datasets used in our paper are all publicly available and are widely adopted by researchers, utilizing LLMs for data annotation and generating few-shot examples may include bias and unfairness. Allowing LLMs to output multiple annotations may further amplify such issues, although we did not observe such phenomena in our experiments. Nevertheless, if CanDist is used with such biased annotations, it may unpleasantly yield unfair and biased predictions based on characteristics like race, gender, disabilities, LGBTQ, or political orientation. To alleviate this issue, we recommend that potential users first use bias reduction and correction techniques to remove biased text and predictions so as to improve overall fairness and ethical standards.

\section*{Acknowledgments}
Haobo Wang is supported by the NSFC under Grants (No. 62402424) and (No. U24A201401).

\bibliography{main}

\appendix

\section{Additional Experimental Setup}

\subsection{Statistics of Datasets}

\begin{table}[ht]
    \caption{Statistics of the used datasets. \textbf{\#Class} denotes the number of classes. \textbf{\#Train} and \textbf{\#Test} indicate the size of the training and testing set.}
    \label{tab:datasets}
    \centering
    \renewcommand\arraystretch{1.05}
    \scalebox{0.85}{
    \begin{tabular}{lcc|cc}
    \toprule
        \textbf{Dataset} & \textbf{Task} & \textbf{\#Class} & \textbf{\#Train} & \textbf{\#Test} \\ 
    \midrule
        TREC & Topic cls & 6 & 5,452 & 500 \\ 
        MA & Medical cls & 5 & 11,550 & 2,888 \\ 
        DBpedia & Ontology  cls & 14 & 10,000 & 70,000 \\ 
        AGNews & Topic cls & 4 & 10,000 & 7,600 \\
        RCT & Content cls & 5 & 10,000 & 30,135 \\
        Banking & Intent cls & 77 & 9,003 & 3,080 \\   
    \bottomrule
    \end{tabular}}
\end{table}

\begin{table*}[!t]
    \caption{Comparisons on $1-\alpha$-error, average number of labels (\#La.), and F1-score between different prompts.}
    \label{tab:assessment}
    \centering
    \renewcommand\arraystretch{1.1}
    \scalebox{0.68}{
    \begin{tabular}{l|ccc|ccc|ccc|ccc|ccc|ccc}
    \toprule
        \multirow{2}{*}{Method} & \multicolumn{3}{c|}{\textbf{TREC}} & \multicolumn{3}{c|}{\textbf{MA}} & \multicolumn{3}{c|}{\textbf{BANK}} & \multicolumn{3}{c|}{\textbf{AGN}} & \multicolumn{3}{c|}{\textbf{RCT}} & \multicolumn{3}{c}{\textbf{DBP}} \\ 
    \cmidrule(r){2-19}
        ~ & $1-\alpha$ & \#La. & F1 & $1-\alpha$ & \#La. & F1 & $1-\alpha$ & \#La. & F1 & $1-\alpha$ & \#La. & F1 & $1-\alpha$ & \#La. & F1 & $1-\alpha$ & \#La. & F1 \\ 
    \midrule
        % Zero-shot & 62.84 & 1.00  & 77.2 & 62.03 & 1.00  & 76.6 & 65.19 & 1.00  & 79.0 & 87.72 & 1.00  & \underline{93.5} & 93.33 & 1.00  & 96.6 \\ 
        SA & 71.07 & 1.00  & 83.1 & 62.28 & 1.00  & 76.8 & 66.08 & 1.00  & 79.6 & 88.73 & 1.00  & \textbf{94.0} & 65.18 & 1.00 & 78.92 & 95.41 & 1.00  & \underline{97.7} \\ 
        $\text{CA}_\text{add}$ & \underline{74.65} & 1.07  & \underline{85.0} & \underline{79.06} & 1.56  & \textbf{82.4} & \underline{76.99} & 1.74  & \underline{86.6} & \underline{94.47} & 1.30  & 92.2 & \underline{75.18} & 1.56 & \textbf{80.26} & \underline{98.59} & 1.37  & \textbf{97.9} \\ 
        $\text{CA}_\text{all}$ & \textbf{89.09} & 1.70  & \textbf{87.5} & \textbf{88.99} & 1.95  & \underline{82.1} & \textbf{80.14} & 2.00  & \textbf{88.5} & \textbf{97.19} & 1.70  & 85.7 & \textbf{79.15} & 1.81 & \underline{79.51} & \textbf{99.25} & 1.75  & 96.7 \\ 
    \bottomrule
    \end{tabular}}
\end{table*}

Table \ref{tab:datasets} shows the statistics of datasets used in our experiments. Given the extensive size of the original training sets for DBpedia, AGNews, and RCT, we randomly selected 10,000 examples from each as their respective training sets. Note that the most competitive baseline, FreeAL, primarily evaluates binary classification datasets, which are easier to annotate and do not need to apply candidate annotations, whereas we conduct experiments on more challenging tasks.

\subsection{More Details of SLM Distillation}

During SLM distillation, we incorporate consistency regularization and mixup training to boost performance following FreeAL.
Consistency regularization encourages the model to produce similar outputs for different augmented views of the same sample. Specifically, we adopt back-translation \citep{DBLP:conf/acl/SennrichHB16} to augment each sample $\bm{x}_i$ into $\bm{x}_i^\text{aug}$. Then, for samples in $\mathcal{D}_\text{in}$ and $\mathcal{D}_\text{out}$, the consistency regularization are formulated as:
\begin{equation}
\begin{aligned}
    \mathcal{L}_\text{cr}^\text{in} &= \frac{1}{|\mathcal{D}_\text{in}|} \sum_{\bm{x}_i \in \mathcal{D}_\text{in}} l_\text{ce}(\bm{p}_i^\text{aug}, \hat{\bm{q}}_i) \\
    \mathcal{L}_\text{cr}^\text{out} &= \frac{1}{|\mathcal{D}_\text{out}|} \sum_{\bm{x}_i \in \mathcal{D}_\text{out}} l_\text{kl}(\bm{p}_i^\text{aug}, \bm{p}_i) \\
\end{aligned}
\end{equation}
where $l_\text{kl}$ denotes the KL-divergence. For mixup training, we create virtual training samples by linearly interpolating both:
\begin{equation}
\begin{aligned}
    \bm{g}(\bm{x}^m) &= \omega\cdot\bm{g}(\bm{x}_i) + (1 - \omega)\cdot\bm{g}(\bm{x}_j) \\
    \hat{\bm{q}}^m &= \omega\cdot\hat{\bm{q}}_i + (1 - \omega)\cdot\hat{\bm{q}}_j
\end{aligned}
\end{equation}
where $\bm{g}(\bm{x}_i)$ is the embedding of $\bm{x}_i$. $\omega \sim \text{Beta}(\varsigma,\varsigma)$ where $\varsigma$ is simply set as 4. The mixup loss $\mathcal{L}_\text{mix}$ is then defined by the cross-entropy loss between the SLM's prediction on $\bm{g}(\bm{x}^m)$ and $\bm{y}_m$. The total loss for SLM distillation is aggregated as:
\begin{equation}
    \mathcal{L}_\text{total} = \mathcal{L}_\text{dr} + \eta\cdot(\mathcal{L}_\text{cr}^\text{in} + \mathcal{L}_\text{cr}^\text{out} + \mathcal{L}_\text{mix})
\end{equation}

\subsection{More Implementation Details}\label{subsec:details}
In our main experiments, we use the gpt-3.5-turbo-0125 version for the LLM API. For generating few-shot examples, we follow the setting in FreeAL which first queries the LLM to generate an example pool of size 100 with corresponding labels. Then, the few-shot examples for each unlabeled sample are retrieved based on embedding similarity with the bert-base-uncased model.

For SLM distillation, we use Nvidia RTX A5000 GPU to train the model for 50 epochs with AdamW optimizer with a learning rate selected from $\{3e-5, 1e-5, 3e-6\}$ and a weight decay of 0.01. The batch size is fixed as 32 with a maximum sequence length of 128. 
We warm up the model by training on the re-normalized distribution for a few epochs to achieve high-quality selection in the Distribution Refinery mechanism.
For hyper-parameters, the small loss ratio $\delta$ is selected from $\{0.4, 0.5, 0.6\}$. The sharpen parameter $\gamma$ is fixed as 0.85 and the high confidence threshold is selected from $\{0.95, 0.99, 1.0\}$.
Note that we employ the default validation set for each dataset for parameter selection.
The loss weight parameter $\eta$ is linearly ramped up from 0 to 1 to avoid overfitting false labels at the start.

\section{Additional Experimental Results}

\subsection{Full Assessment Results}\label{subsec:full_assessment}
In this section, we demonstrate the assessment results of single annotations and candidate annotations on all tasks (training sets), where we use the average number of labels (\#La.) to represent $\beta$-coverage since it is more intuitive to understand. As shown in Table \ref{tab:assessment}, $\text{CA}_\text{add}$ and $\text{CA}_\text{all}$ improve $1-\alpha$-error on all datasets with average number of labels no more than two. Candidate annotations also achieve higher F1-scores on all tasks except for AGNews. These results statistically demonstrate that candidate annotations are more likely to include the correct labels and offer great potential.

\begin{table}[!t]
    \caption{Assessment results of different prompting strategies on TREC using Llama 3.1 and GPT-4o.}
    \label{tab:assessment_4o}
    \centering
    \renewcommand\arraystretch{1.1}
    \scalebox{0.83}{
    \begin{tabular}{l|ccc|ccc}
    \toprule
        \multirow{2}{*}{Method} & \multicolumn{3}{c|}{\textbf{Llama 3.1}} & \multicolumn{3}{c}{\textbf{GPT-4o}} \\ 
    \cmidrule(r){2-7}
        ~ & $1-\alpha$ & \#La. & F1 & $1-\alpha$ & \#La. & F1 \\ 
    \midrule
        SA & 68.80 & 1.00  & 81.52 & 87.53 & 1.00  & 93.35  \\ 
        $\text{CA}_\text{add}$ & \underline{85.34} & 1.87  & \textbf{83.98} & \underline{94.42} & 1.20  & \textbf{95.17} \\ 
        $\text{CA}_\text{all}$ & \textbf{89.56} & 2.06  & \underline{83.80} & \textbf{96.28} & 1.44  & \underline{93.63} \\ 
    \bottomrule
    \end{tabular}}
\end{table}

\begin{table}[!t]
    \caption{Comparisons on the training set and testing set of TREC using Llama 3.1 and GPT-4o.}
    \label{tab:accuracy_4o}
    \centering
    \renewcommand\arraystretch{1.05}
    \scalebox{0.85}{
    \begin{tabular}{l|p{1.11cm}<{\centering}p{1.11cm}<{\centering}|p{1.11cm}<{\centering}p{1.11cm}<{\centering}}
    \toprule
        \multirow{2}{*}{Method} & \multicolumn{2}{c|}{\textbf{Llama 3.1}} & \multicolumn{2}{c}{\textbf{GPT-4o}} \\ 
    \cmidrule(r){2-5}
        ~ & Train & Test & Train & Test \\ 
    \midrule
        Few-shot & 68.80 & 77.00 & 87.53 & 87.60 \\ 
        FreeAL & 76.60 & 82.67 & 89.14 & 93.80 \\ 
        $\text{CanDist}_\text{add}$ & \underline{76.99} & \underline{83.40} & \underline{89.53} & \underline{95.60} \\ 
        $\text{CanDist}_\text{all}$ & \textbf{77.66} & \textbf{85.60} & \textbf{90.48} & \textbf{96.40} \\ 
    \bottomrule
    \end{tabular}}
\end{table}

\subsection{Results of Different LLMs}\label{subsec:4o}
In this section, we evaluate the annotation results using two other LLMs: Llama 3.1 (Llama-3.1-8B-Instruct) and GPT-4o. As shown in Table \ref{tab:assessment_4o} and \ref{tab:accuracy_4o}, Llama 3.1 achieves results at the same level as GPT-3.5, and using the more advanced GPT-4o boosts the performance of all data annotation methods. Still, CanDist improves GPT-4o's single annotations (Few-shot) by a large margin of \textbf{8.80\%} and outperforms the most competitive baseline FreeAL by a margin of \textbf{2.60\%} on the testing set.

\subsection{Synergism with Self-Consistency}\label{subsec:mv}
\begin{figure}[!t]
    \centering
    \includegraphics[width=0.99\linewidth]{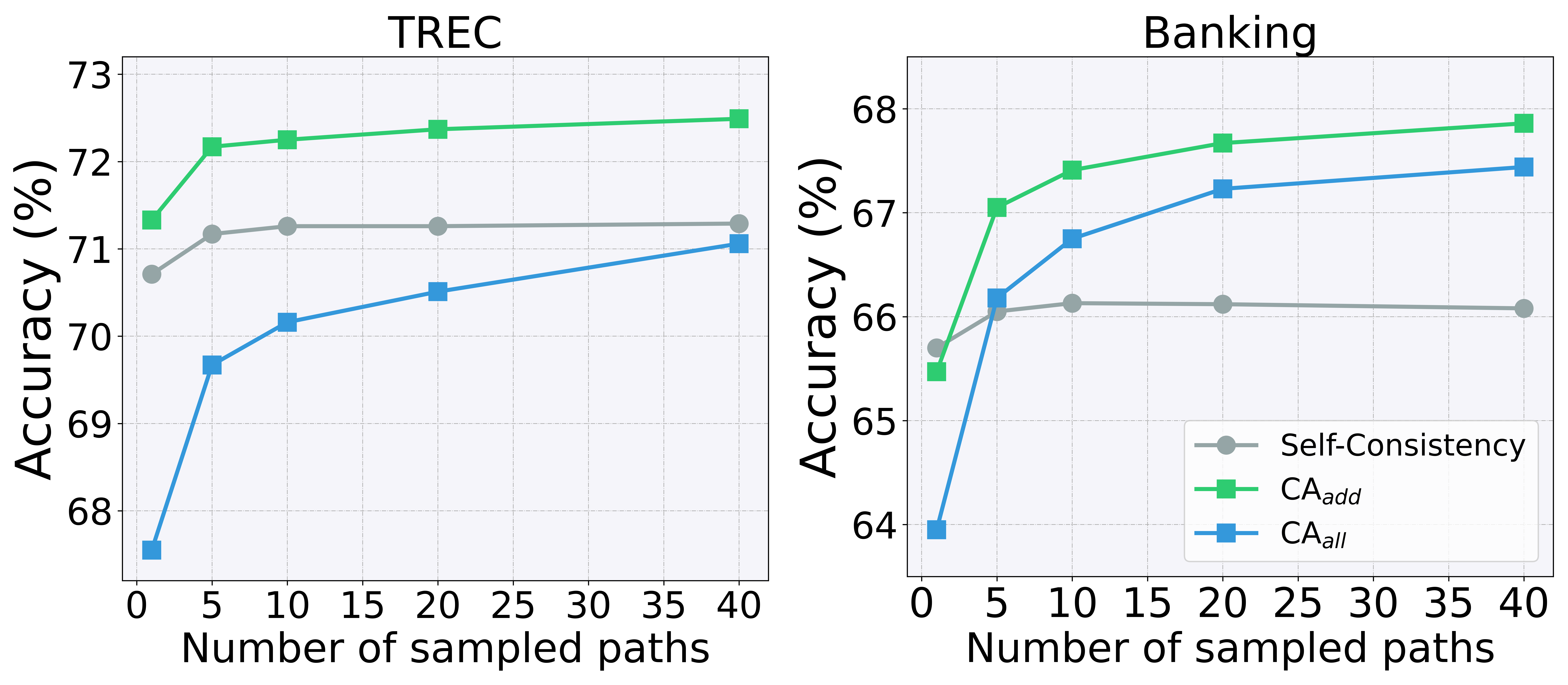}
    \caption{Comparison of different prompting strategies for self-consistency shows the synergism between prompting candidates with self-consistency.}
    \label{fig:ablation_mv}
\end{figure}

Following the setting in paragraph \ref{para:scmv}, we further show that the collaboration of prompting candidates and majority voting (i.e. SC-1) also brings great potential by outperforming voting on single annotations. Specifically, after sampling $K=40$ candidate annotations, we use majority voting to obtain the final annotation: $\hat{y}=\argmax_{c\ \in \mathcal{Y}}\sum_{j=1}^K \mathbb{I}(c \in s_j)$. Figure \ref{fig:ablation_mv} demonstrates the comparison results on the training set of TREC and Banking, where we found that voting on candidate annotations results in higher performance than voting on single annotations. Notably, as the number of sampled paths increases, the accuracy of voting on candidates grows more significantly, especially from 1 to 5. This further indicates the great value of prompting candidate annotations.

\subsection{Comparison of Different ICL Strategies for Prompting Candidates}
In this section, we further investigate how the design of in-context learning (ICL) examples for prompting candidate annotations affects the results of CanDist. Note that we employ \textit{Self-generated (Single)} for our method following FreeAL, which leverages sample-single label pairs generated by LLM as ICL examples. We further explore the effect of two other types of ICL examples: \textit{Self-generated (Candidate)} which leverages sample-candidate label pairs generated by LLM as examples; \textit{Supervised} adopt human-labeled training data as examples. For both methods, we first gather an example pool of size 100 and retrieve ICL examples for each unlabeled sample based on embedding similarity with the bert-base-uncased model. As shown in Table \ref{tab:icl}, CanDist using self-generated examples outperforms zero-shot CanDist, and using supervised ICL can make further improvements. Besides, CanDist using examples with self-generated single labels outperforms the one with candidate labels on Banking but underperforms it on TREC. This suggests that whether to use single labels or candidate labels as ICL examples depends on the specific task and we simply adopt the former, which achieves state-of-the-art results.

\begin{table}[!t]
    \caption{Comparison of different ICL strategies for prompting candidate annotations.}
    \label{tab:icl}
    \centering
    \renewcommand\arraystretch{1.05}
    \scalebox{0.90}{
    \begin{tabular}{l|cc}
    \toprule
        \textbf{Example Type} & \textbf{TREC} & \textbf{BANK} \\ 
    \midrule
        Zero-shot & 87.00 & 68.47  \\ 
        Self-generated (Single) & 87.80 & \underline{75.97}  \\ 
        Self-generated (Candidate) & \underline{89.60} & 74.71  \\ 
        Supervised & \textbf{90.47} & \textbf{76.04}  \\
    \bottomrule
    \end{tabular}}
\end{table}

\begin{figure}[!t]
    \centering
    \includegraphics[width=0.40\textwidth]{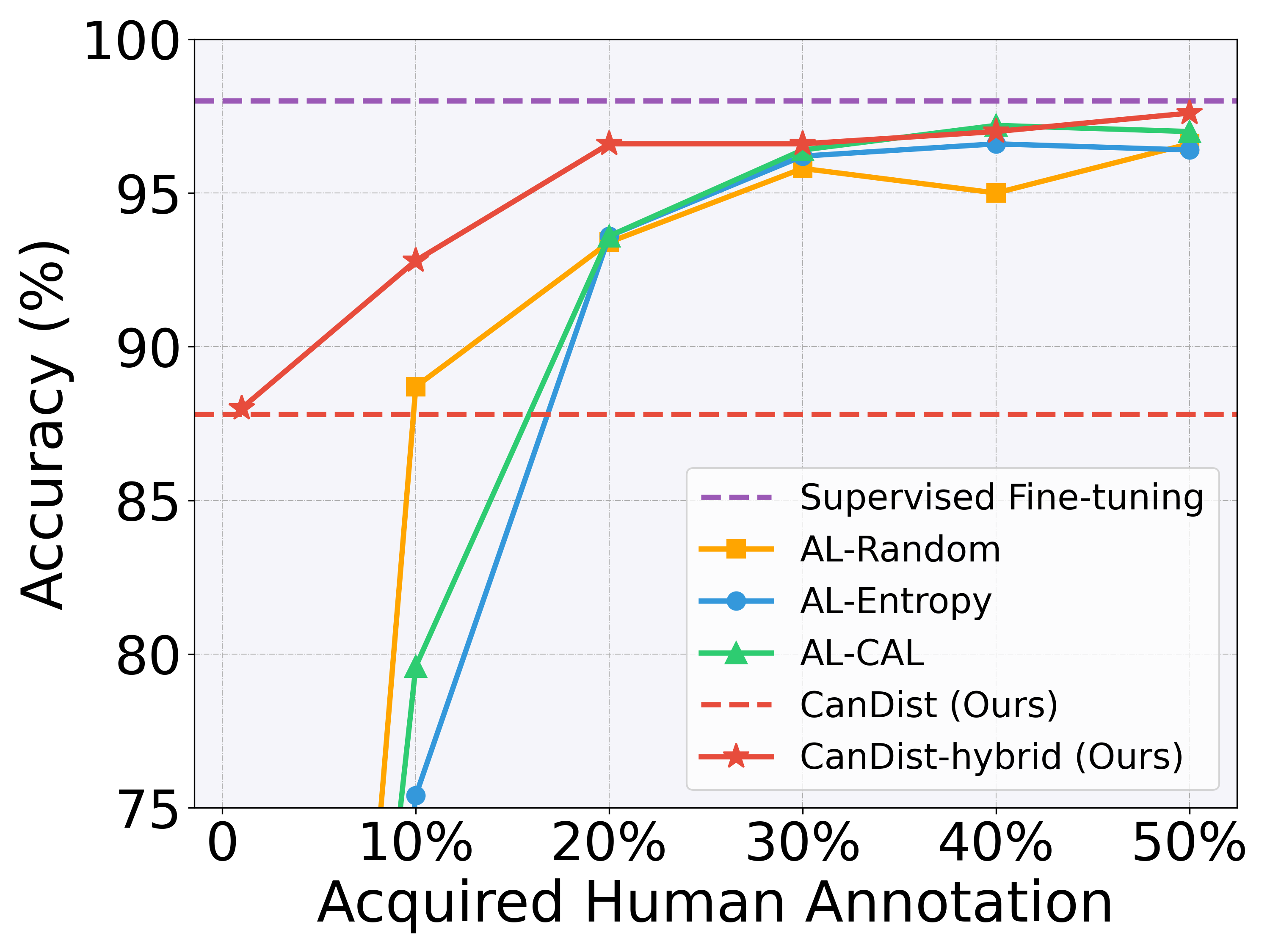}
    \caption{Comparison between active learning methods and CanDist on TREC where $\text{CanDist}_\text{all}$ is applied.}
    \label{fig:al}
\end{figure}

\subsection{Comparison with Traditional Active Learning Methods}
To compare the effectiveness of CanDist with human annotation, we further evaluate some active learning (AL) baselines, including 1) \textit{AL-Random}, which acquires to-be-labeled data randomly; 2) \textit{AL-Entropy} \citep{DBLP:conf/cvpr/HolubPB08}, which is the most commonly used uncertainty-based method that acquires samples with highest predictive entropy; 3) \textit{AL-CAL} \citep{DBLP:conf/emnlp/MargatinaVBA21} is a recent active learning method that acquires contrastive examples. We also report \textit{Supervised Fine-tuning} which acquires annotation for the whole training set and \textit{CanDist-hybrid} which incorporates randomly acquired human annotations into CanDist. For all methods, we first train the SLM on the annotated training set and evaluate its testing accuracy.

Figure \ref{fig:al} demonstrates the comparison results under different annotation budgets on the TREC datasets. Firstly, CanDist, without human annotation, outperforms most traditional AL baselines under 10\% human annotations. Also, incorporating merely 20\% human annotations, CanDist-hybrid achieves comparable performance with AL baselines under 50\% human annotations. Furthermore, CanDist-hybrid with 50\% human annotations achieves competitive performance on par with supervised fine-tuning. These results yield the superiority of our proposed CanDist framework.

Besides, though FreeAL shows that LLM-driven active learning surpasses traditional active learning and achieves competitive results with supervised fine-tuning on the SST-2 \citep{DBLP:conf/emnlp/SocherPWCMNP13} and MR \citep{DBLP:conf/acl/PangL05} datasets, we show that on a harder task, LLM-driven active learning still requires a small proportion of human annotations to achieve near-supervised performance.

\begin{table}[!t]
    \caption{Running time (in seconds) of one SLM training epoch of baseline FreeAL and CanDist.}
    \label{tab:time}
    \centering
    \renewcommand\arraystretch{1.15}
    \scalebox{0.77}{
    \begin{tabular}{l|cccccc}
    \toprule
        \textbf{Method} & \textbf{TREC} & \textbf{MA} & \textbf{DBP} &  \textbf{AGN} & \textbf{RCT} & \textbf{BANK} \\ 
    \midrule
        FreeAL & 80.2 & 172.7 & 148.5 & 147.8 & 146.2 & 131.7 \\ 
        CanDist & 79.1 & 174.0 & 149.4 & 148.1 & 146.5 & 132.4 \\
    \bottomrule
    \end{tabular}}
\end{table}

\subsection{Time Complexity Analysis}
To analyze the time complexity of the SLM distillation process in our proposed CanDist, we compare the empirical running time (in seconds) of SLM distillation in CanDist and the baseline FreeAL in Table \ref{tab:time}, which demonstrates CanDist is in the same magnitude as FreeAL.

\section{Proof of Theorem \ref{thm}}\label{sec:proof}

In this section, we provide the proof of Theorem \ref{thm}, which illustrates that the SLM distilled from the LLM's candidate annotations enjoys better theoretical guarantees than the LLM as well as the SLM distilled from the LLM's single annotations.

\setcounter{theorem}{0} 
\begin{theorem}
\label{app:thm}
Considering the scenario that both the teacher LLM and student SLM are composed of a feature extractor $\bm{g}(\cdot):\mathcal{X}\mapsto \mathbb{R}^d$ (with different scales) and a classifier $\bm{W}\in\mathbb{R}^{d\times C}$. The teacher LLM is pre-trained on an inaccurate dataset $\tilde{\mathcal{D}}=\{\bm{x}_i, \tilde{y}_i \}_{i=1}^m$ with noise rates $\{\bm{R}_{c, c'}\}_{c=1, c'=1}^{C,C}$, where $m$ denotes the number of samples in the dataset and $\bm{R}_{c, c'}$ indicates the probability of label $c$ being flipped to $c'$. After pre-training, the student SLM is then trained based on the teacher LLM's single (top-1) or candidate (top-2) annotations on $\tilde{\mathcal{D}}$. Suppose the models are trained by $l_2$-regularized cross-entropy loss with regularization parameter $\lambda$, and the feature extractors are fixed. Besides, we consider that the feature similarity between different samples from the same class and different classes are $a$ and $b$ respectively, with $1>a>b>0$.

Then, with $m\to \infty$, the condition of achieving 100\% accuracy (correctly predicting all training data) for the teacher LLM as well as the student SLM distilled from LLM's top-1 prediction is:
\begin{equation}
\label{eq:p_after1}
\begin{gathered}
    \bm{R}_{c, c'} + \sum_{i\neq c}\bm{R}_{c, i} < 1 - \frac{\theta}{\phi-\theta},~\forall c, c'\neq c \\ 
    \text{where} ~ \theta = 1-\frac{Cm\lambda}{Cm\lambda+1-a}, \\
    \phi = 1 - \frac{Cm\lambda}{Cm\lambda + \frac{m}{C}(a-b)+1-a} \\
\end{gathered} 
\end{equation}
and the condition of that for the student SLM distilled from LLM's top-2 prediction is:
\begin{equation}
\label{eq:p_after2}
    \bm{R}_{c, c'} + \sum_{i\neq c}\bm{R}_{c, i} < 1 ,~\forall c,c'\neq c
\end{equation}
\end{theorem}

\noindent\textit{Proof}.
\paragraph{Closed-form Solutions of Model's Prediction.} Denote the training objective of the models as:
\begin{equation}
    \mathcal{L}(\bm{W})=\frac{1}{m}\sum_{i=1}^{m} l_\text{ce}(\bm{p}_i, \bm{q}_i) + \frac{\lambda ||\bm{W}||_F^2}{2}
\end{equation}
where $\bm{p}_i=\mathbf{softmax}(\bm{W}^\top\bm{g}(\bm{x}_i))$ is the model's prediction distribution and $\bm{q}_i$ denotes the training target. When pre-training the teacher LLM, $\bm{q}_i = \mathbf{e}(\tilde{y}_i)$ where $\mathbf{e}(y)$ denotes the one-hot form of a specific label $y$; When distilling the student SLM from teacher LLM's top-1 prediction, $\bm{q}_i$ is a one-hot vector where the value on the max prediction index equals 1 and otherwise 0; When distilling the student SLM from teacher LLM's top-2 prediction, $\bm{q}_i$ is a vector where the value on the top-2 prediction index equals 0.5 and otherwise 0.

The optimal classifier satisfies the condition of $\frac{d\mathcal{L}(\bm{W})}{d\bm{W}} = \frac{1}{m}\sum_{i=1}^m \bm{g}(\bm{x}_i)(\bm{p}_i-\bm{q}_i)^\top + \lambda\bm{W}=0$. Thus, the optimal classifier can be formalized as:
\begin{equation}
    \bm{W}^\top = \frac{1}{m\lambda} \sum_{i=1}^m (\bm{q}_i - \bm{p}_i)\bm{g}(\bm{x}_i)^\top
\end{equation}

To derive the relation between the training target $\bm{q}_i$ and model's prediction $\bm{p}_i$, we define $\bm{a}_i=\bm{q}_i - \bm{p}_i$ and derive as follows:
\begin{equation*}
\begin{aligned}
    \bm{a}_i &= \bm{q}_i - \bm{p}_i = \bm{q}_i - \mathbf{softmax}(\bm{W}^\top\bm{g}(\bm{x}_i)) \\
    &= \bm{q}_i - \mathbf{softmax}(\frac{1}{m\lambda} \sum_{j=1}^m \bm{a}_j\bm{g}(\bm{x}_j)^\top\bm{g}(\bm{x}_i)) \\
    &= \bm{q}_i - \mathbf{softmax}(\frac{1}{m\lambda} \sum_{j=1}^m \langle\bm{g}(\bm{x}_i), \bm{g}(\bm{x}_j)\rangle \bm{a}_j)
\end{aligned}
\end{equation*}
Due to the non-linearity of the softmax function, directly solving $\bm{a}_i$ is challenging. To this end, we employ a linear approximation of the softmax function following \citep{DBLP:journals/corr/HintonVD15}:
\begin{equation}
\label{eq:p_linear}
\begin{aligned}
    \mathbf{softmax}(\bm{v})_i &= \frac{\exp(v_i)}{\sum_{j=1}^C \exp(v_j)} \\
    &\approx \frac{1+v_i}{C+\sum_{j=1}^C v_j} \approx \frac{1+v_i}{C}
\end{aligned}
\end{equation}
Note that this linear approximation, originally introduced by \citet{DBLP:journals/corr/HintonVD15}, is based on applying softmax with a high temperature $T>0$, i.e., $\mathbf{softmax}(\bm{v}/T)$. Therefore, when $T=1$, the approximation in Eq.(\ref{eq:p_linear}) becomes valid when the logits are of sufficiently small magnitude. By applying the above approximation, we have:
\begin{equation}
\label{eq:p_vector}
    \bm{a}_i = \bm{q}_i - \frac{1}{C}\mathbf{1}_C - \frac{1}{Cm\lambda}\sum_{j=1}^m \langle\bm{g}(\bm{x}_i), \bm{g}(\bm{x}_j)\rangle \bm{a}_j
\end{equation}
where $\mathbf{1}_C$ a $C$-dimensional all-ones vector. Denoting $\bm{A}=\left[\bm{a}_1, \dots, \bm{a}_m\right]\in\mathbb{R}^{C\times m}$, $\bm{Q}=\left[\bm{q}_1, \dots, \bm{q}_m\right]\in\mathbb{R}^{C\times m}$, and $\bm{S}\in\mathbb{R}^{m\times m}$ with $\bm{S}_{i, j}=\langle\bm{g}(\bm{x}_i), \bm{g}(\bm{x}_j)\rangle$, Eq.(\ref{eq:p_vector}) can be expressed as:
\begin{equation}
    \bm{A} = \bm{Q} - \frac{1}{C}\mathbf{1}_{C\times m} - \frac{1}{Cm\lambda}\bm{A}\bm{S}^\top
\end{equation}
With the definition of $\bm{A}$ and the symmetry of $\bm{S}$, and denote $\bm{P}=\left[\bm{p}_1, \dots, \bm{p}_m\right]\in\mathbb{R}^{C\times m}$ as the output matrix, the relation between the training target $\bm{Q}$ and the model's prediction $\bm{P}$ can be derived as:
\begin{equation}
\label{eq:p_matrix}
\begin{gathered}
    \bm{A} = \bm{Q} - \frac{1}{C}\mathbf{1}_{C\times m} - \frac{1}{Cm\lambda}\bm{A}\bm{S}; \\
    \bm{A}\left(\bm{I}_m+\frac{1}{Cm\lambda}\bm{S}\right) = \bm{Q} - \frac{1}{C}\mathbf{1}_{C\times m}; \\
    \bm{A} = \left(\bm{Q} - \frac{1}{C}\mathbf{1}_{C\times m}\right)\left(\bm{I}_m + \frac{1}{Cm\lambda}\bm{S}\right)^{-1}; \\
    \left(\bm{Q} - \frac{1}{C}\mathbf{1}_{C\times m}\right) - \left(\bm{P} - \frac{1}{C}\mathbf{1}_{C\times m}\right)= \\
    \quad\quad \left(\bm{Q} - \frac{1}{C}\mathbf{1}_{C\times m}\right)\left(\bm{I}_m + \frac{1}{Cm\lambda}\bm{S}\right)^{-1}; \\
    \bm{P} - \frac{1}{C}\mathbf{1}_{C\times m}= \left(\bm{Q} - \frac{1}{C}\mathbf{1}_{C\times m}\right) \\
    \quad\quad\left(\bm{I}_m - \left(\bm{I}_m + \frac{1}{Cm\lambda}\bm{S}\right)^{-1}\right)
\end{gathered}
\end{equation}
where $\bm{I}_m$ is an $m$-dimensional identity matrix. To further simplify the above expression, we apply eigen-decomposition for the similarity matrix $\bm{S}$ as $\bm{S}=\bm{V}\Lambda\bm{V}^{-1}$ with eigenvalue-eigenvector pairs $\{\lambda_i, \bm{v}_i\}_{i=1}^m$. Then, by applying Woodbury’s matrix identity, Eq.(\ref{eq:p_matrix}) can be simplified as:
\begin{equation}
\label{eq:p_wood}
\begin{gathered}
    \bm{P} - \frac{1}{C}\mathbf{1}_{C\times m}=\left(\bm{Q} - \frac{1}{C}\mathbf{1}_{C\times m}\right) \\
    \quad\quad\left(\bm{I}_m - \left(\bm{I}_m + \bm{V}\frac{1}{Cm\lambda}\Lambda\bm{V}^{-1}\right)^{-1}\right) \\
    = \left(\bm{Q} - \frac{1}{C}\mathbf{1}_{C\times m}\right)\bm{V}\left(Cm\lambda\Lambda^{-1}+\bm{I}_m\right)\bm{V}^{-1} \\
    % = \left(\bm{Q} - \frac{1}{C}\mathbf{1}_{C\times m}\right)\sum_{i=1}^m \frac{\lambda_i}{Cm\lambda + \lambda_i}\bm{v}_i\bm{v}_i^\top
\end{gathered}
\end{equation}

\paragraph{Quantification of the Similarity Matrix.} In the following derivations, we further simplify the closed-form solution of $\bm{P}$ through the quantification of the similarity matrix $\bm{S}$. Specifically, we assume that the feature similarity of different samples depends on classes, i.e.:
\begin{equation}
    \begin{aligned}
       \bm{S}_{i, j} = \begin{cases}
        1, & i=j \\
        a, & i\neq j, y_i = y_j \\
        b, & y_i \neq y_j \\
    \end{cases}
    , \text{where}~b<a<1
    \end{aligned}
\end{equation}
Denote the class-wise similarity matrix $\bm{Z}\in\mathbb{R}^{C\times C}$ with $\bm{Z}_{i, j}=a$ when $i=j$ and $\bm{Z}_{i, j}=b$ when $i\neq j$, and let $\bm{Y}=\left[\mathbf{e}(y_1), \dots, \mathbf{e}(y_m)\right]\in\mathbb{R}^{C\times m}$ be the ground-truth label matrix, the similarity matrix $\bm{S}$ can be expressed as:
\begin{equation}
\label{eq:identity_shift}
\begin{aligned}
    \bm{S} &= \bm{Y}^\top\bm{Z}\bm{Y} + (1-a)\bm{I}_m \\
    &= \bm{Y}^\top\left(b\mathbf{1}_{C\times C}+\left(a-b\right)\bm{I}_C\right)\bm{Y} + (1-a)\bm{I}_m
\end{aligned}
\end{equation}

\setcounter{theorem}{0}
\begin{lemma}
\label{lma:1}
Suppose the symmetric matrix $\bm{B}\in\mathbb{R}^{n\times n}$ is composed of the sum of rank-$m$ ($m<n$) matrix and a multiple of the identity matrices:
\begin{equation*}
    \bm{B} = \bm{U}\Xi\bm{U}^\top + \lambda\bm{I}_n
\end{equation*}
where $\bm{U}=\left[\bm{u}_1, \dots, \bm{u}_m \right]\in\mathbb{R}^{n\times m}$ is an orthonormal matrix satisfying $\bm{U}^\top \bm{U}=\bm{I}_m$. $\Xi=diag(\xi_1, \dots, \xi_m)\in\mathbb{R}^{m\times m}$ containing the eigenvalues $\xi_i$. Then, $\bm{B}$ has the following two types of eigenvalue-eigenvector pairs $\{\sigma_i, \bm{v}_i\}_{i=1}^{n}$: 

1) $m$ eigenvalues that are shifts of the original eigenvalues from the rank-$m$ matrix:
\begin{equation*}
    \sigma_i = \xi_i + \lambda, \quad i=1, \dots, m
\end{equation*}
with corresponding eigenvectors $\bm{v}_i=\bm{u}_i$.

2) $(n-m)$ eigenvalues from the identity matrix:
\begin{equation*}
    \sigma_i = \lambda, \quad i=m+1, \dots, n
\end{equation*}
with corresponding eigenvectors orthogonal to the columns of $\bm{U}$.
\end{lemma}
\noindent\textit{Proof}. The eigenvalue equation is given by:
\begin{equation*}
    (\bm{U}\Xi\bm{U}^\top + \lambda\bm{I}_n)\bm{v}=\sigma \bm{v}
\end{equation*}
Decompose $\bm{v}$ into components $\bm{v}_\parallel + \bm{v}_\perp$, where $\bm{v}_\parallel$ is in the column space of $\bm{U}$ and $\bm{v}_\perp$ is orthogonal to the column space of $\bm{U}$, and we have $\bm{v}_\parallel=\bm{U}\bm{\beta}$ and $\bm{U}^\top\bm{v}_\perp=\mathbf{0}$ for some $\bm{\beta}\in\mathbb{R}^m$. Then, multiplying $\bm{U}^\top$ on both sides of the eigenvalue equation yields:
\begin{equation*}
\begin{gathered}
    \Xi\bm{U}^\top\bm{v} + \lambda\bm{U}^\top\bm{v} = \sigma\bm{U}^\top\bm{v}; \\
    \Xi\bm{U}^\top(\bm{v}_\parallel + \bm{v}_\perp) + \lambda\bm{U}^\top(\bm{v}_\parallel + \bm{v}_\perp) \\
    \quad\quad=\sigma\bm{U}^\top(\bm{v}_\parallel + \bm{v}_\perp); \\
    (\Xi + \lambda\bm{U}^\top\bm{U})\bm{\beta}=\sigma\bm{\beta}; \\
    (\Xi + \lambda\bm{I})\bm{\beta}=\sigma\bm{\beta}
\end{gathered}
\end{equation*}
which indicates $\sigma_i=\xi_i + \lambda$ for $i=1,\dots,m$ with corresponding eigenvectors given by $\bm{v}_i=\bm{u}_i$. The remaining $n-m$ eigenvalues arise from $\lambda\bm{I}$, with eigenvectors orthogonal to the columns of $\bm{U}$.

With Lemma \ref{lma:1}, we can reformulate $\bm{S}$ in Eq.(\ref{eq:identity_shift}). For $\bm{Z} = b\mathbf{1}_{C\times C}+\left(a-b\right)\bm{I}_C$, it has two types of eigenvalue-eigenvector pairs $\{\sigma_i, \bm{u}_i\}_{i=1}^C$: 

1) one pair with eigenvalue:
\begin{equation*}
    \sigma_1=Cb + (a-b)
\end{equation*}
and eigenvector $\bm{u}_1=\frac{1}{\sqrt{C}}\mathbf{1}_C$;

2) $C-1$ pairs with eigenvalues:
\begin{equation*}
    \sigma_i=a-b,\quad i=2,\dots,C
\end{equation*}
and the corresponding eigenvectors $\bm{u}_i$. Denoting $\Sigma=diag(\sigma_1, \dots, \sigma_C)$ and $\bm{U}=\left[\bm{u}_1, \dots, \bm{u}_C\right]\in\mathbb{R}^{m\times C}$, thus:
\begin{equation}
\begin{aligned}
    \bm{S} &= \bm{Y}^\top\bm{Z}\bm{Y} + (1-a)\bm{I}_m \\
    &= \bm{Y}^\top\bm{U}\Sigma\bm{U}^\top\bm{Y} + \left(1-a\right)\bm{I}_m \\
    &= \sqrt{\frac{C}{m}}\bm{Y}^\top\bm{U} \left(\frac{m}{C}\Sigma\right) \left(\sqrt{\frac{C}{m}}\bm{Y}^\top\bm{U}\right)^\top \\
    &\quad + \left(1-a\right)\bm{I}_m
\end{aligned}
\end{equation}
where we assume $\sum_{j=1}^m \bm{Y}_{i, j}=m/C$, namely, the dataset is balanced. Again, by applying Lemma \ref{lma:1}, $\bm{S}$ has three types of eigenvalue-eigenvector pairs $\{\lambda_i, \bm{v}_i\}_{i=1}^m$:

1) one pair with eigenvalue:
\begin{equation*}
\begin{aligned}
    \lambda_1 &= \frac{m}{C}\sigma_1 +(1-a) \\
    &= mb+\frac{m}{C}(a-b)+(1-a)
\end{aligned}
\end{equation*}
and eigenvector $\bm{v}_1=\sqrt{\frac{C}{m}}\bm{Y}^\top\bm{u}_1 = \frac{1}{\sqrt{m}}\bm{Y}^\top\mathbf{1}_C$;

2) $C-1$ pairs with eigenvalues for $i=2,\dots,C$:
\begin{equation*}
\begin{aligned}
    \lambda_i &= \frac{m}{C}\sigma_i +(1-a) \\
    &= \frac{m}{C}(a-b) + (1-a)
\end{aligned}
\end{equation*}
and the eigenvectors $\bm{v}_i=\sqrt{\frac{C}{m}}\bm{Y}^\top\bm{u}_i$;

3) $m-C$ pairs with eigenvalues:
\begin{equation*}
    \lambda_i = (1-a), \quad i=C+1,\dots,m
\end{equation*}
and the corresponding eigenvectors $\bm{v}_i$.

Denoting $\bm{S}' = \bm{V}\left(Cm\lambda\Lambda^{-1}+\bm{I}_m\right)\bm{V}^{-1}$ in Eq.(\ref{eq:p_wood}), and denoting $\theta, \phi, \psi$ according to the following equations:
\begin{equation*}
\begin{aligned}
    \theta &= 1 - \frac{Cm\lambda}{Cm\lambda+1-a} \\
    \phi &= 1 - \frac{Cm\lambda}{Cm\lambda + \frac{m}{C}(a-b)+1-a} \\
    \psi &= 1 - \frac{Cm\lambda}{Cm\lambda + mb+\frac{m}{C}(a-b)+1-a}
\end{aligned}
\end{equation*}
we have:
\begin{equation}
\begin{aligned}
    \bm{S}' &= \sum_{i=1}^m \frac{\lambda_i}{Cm\lambda + \lambda_i}\bm{v}_i\bm{v}_i^\top \\
    &= \frac{\lambda_1}{Cm\lambda + \lambda_1}\bm{v}_1\bm{v}_1^\top + \sum_{i=2}^C \frac{\lambda_i}{Cm\lambda + \lambda_i}\bm{v}_i\bm{v}_i^\top \\
    &\quad +\sum_{i=C+1}^m \frac{\lambda_i}{Cm\lambda + \lambda_i}\bm{v}_i\bm{v}_i^\top \\
    &= \frac{\psi C}{m}\bm{Y}^\top\bm{u}_1\bm{u}_1^\top\bm{Y} + \frac{\phi C}{m}\sum_{i=2}^C \bm{Y}^\top\bm{u}_i\bm{u}_i^\top\bm{Y} \\
    &\quad+ \theta\sum_{i=C+1}^{m}\bm{v}_i\bm{v}_i^\top \\
    &= \frac{\psi}{m}\bm{Y}^\top\mathbf{1}_{C\times C}\bm{Y} \\
    &\quad+ \frac{\phi C}{m}\bm{Y}^\top \left(\bm{I}_C - \frac{1}{C}\mathbf{1}_{C\times C}\right)\bm{Y} \\
    &\quad+ \theta\left(\bm{I}_m - \frac{C}{m}\sum_{i=1}^C\bm{Y}^\top\bm{u}_i\bm{u}_i^\top\bm{Y}\right) \\
    &= \frac{\psi-\phi}{m}\mathbf{1}_{m\times m} + \frac{(\phi-\theta)C}{m}\bm{Y}^\top\bm{Y} + \theta\bm{I}_m
\end{aligned}
\end{equation}
Finally, the model's prediction $\bm{p}_i$ is quantified as:
\begin{equation}
\label{eq:p_quantified}
\begin{aligned}
    \bm{p}_i &= \left(\bm{Q}-\frac{1}{C}\mathbf{1}_{C\times m}\right)\bm{S}'_{:,i}+\frac{1}{C}\mathbf{1}_C \\
    &= \theta\bm{q}_i + \left(\phi-\theta\right)\left(\frac{C}{m}\sum_{j:y_i=y_j}\bm{q}_j\right) \\
    &\quad+ \left(\psi-\phi\right)\left(\frac{1}{m}\sum_{j=1}^m\bm{q}_j\right) + \left(1-\psi\right)\frac{1}{C}\mathbf{1}_C \\
    &= \theta\bm{q}_i + \left(\phi-\theta\right)\left(\frac{C}{m}\sum_{j:y_j=y_i}\bm{q}_j\right) \\
    &\quad+ \left(1-\phi\right)\frac{1}{C}\mathbf{1}_C \\
\end{aligned}
\end{equation}
where we assume the target $\bm{Q}$ is also balanced which indicates $\frac{1}{m}\sum_{j=1}^m\bm{q}_j=\frac{1}{C}\mathbf{1}_C$.

\paragraph{Condition for Achieving Correct Prediction.} Recall that the teacher model is trained on an inaccurate dataset $\tilde{\mathcal{D}}=\{\bm{x}_i, \tilde{y}_i \}_{i=1}^m$ with noise rates $\{\bm{R}_{c, c'}\}_{c=1, c'=1}^{C,C}$, and we have $\bm{q}_i = \mathbf{e}(\tilde{y}_i)$ when training the teacher model. Then, when $m\to \infty$, the second term in Eq.(\ref{eq:p_quantified}) can be expressed as $\frac{C}{m}\sum_{j:y_j=y_i}\bm{q}_j = \bm{R}_{y_i, :}^\top$, which yields:
\begin{equation}
    \bm{p}_i = \theta\mathbf{e}(\tilde{y}_i) + (\phi-\theta)\bm{R}_{y_i, :}^\top + \frac{(1-\phi)}{C}\mathbf{1}_C
\end{equation}

Then, we aim to find the conditions for the prediction $\bm{p}_i$ to have the maximum value at the true label position $y_i$, indicating a correct prediction. On the one hand, if sample $\bm{x}_i$ is clean, i.e., $y_i = \tilde{y}_i$:
\begin{equation}
    \begin{aligned}
        \left[\bm{p}_i\right]_c = \begin{cases}
        \theta + (\phi-\theta)\bm{R}_{y_i, y_i}, & c = y_i \\
        (\phi-\theta)\bm{R}_{y_i, c}, & c \neq y_i \\
    \end{cases}
    \end{aligned}
\end{equation}
where the condition for  $\argmax_c \left[\bm{p}_i\right]_c = y_i$ is $\bm{R}_{c,c}>\bm{R}_{c,c'}-\frac{\theta}{\phi-\theta},\forall c, c'\neq c$; On the other hand, if sample $\bm{x}_i$ is noisy, i.e., $y_i \neq \tilde{y}_i$:
\begin{equation}
\label{eq:p_condition}
    \begin{aligned}
        \left[\bm{p}_i\right]_c = \begin{cases}
        (\phi-\theta)\bm{R}_{y_i, y_i}, & c = y_i \\
        \theta + (\phi-\theta)\bm{R}_{y_i, \tilde{y}_i}, & c = \tilde{y}_i \\
        (\phi-\theta)\bm{R}_{y_i, c}, & c \neq y_i, \tilde{y}_i \\
    \end{cases}
    \end{aligned}
\end{equation}
where the condition is $\bm{R}_{c,c}>\bm{R}_{c,c'}+\frac{\theta}{\phi-\theta},\forall c, c'\neq c$. Overall, since we have $\phi>\theta$, the most stringent condition for correct prediction of the teacher LLM is $\bm{R}_{c,c}>\bm{R}_{c,c'}+\frac{\theta}{\phi-\theta},\forall c, c'\neq c$.

Note that if $\bm{R}_{c,c}<\bm{R}_{c,c'}+\frac{\theta}{\phi-\theta}$ for some $c$ and $c'\neq c$, the teacher model's top-1 prediction on those samples with $y_i=c$ and $\tilde{y}_i=c'$ remains noisy, which indicates that when distilling from the teacher model's top-1 prediction $\bm{Q}$, the noise rates $\{\bm{R}^q\}_{c=1,c'=1}^{C,C}$ for $\bm{Q}$ still satisfies $\bm{R}_{c,c}^{q}<\bm{R}_{c,c'}^{q}+\frac{\theta}{\phi-\theta}$ for those $c$ and $c'\neq c$. To this end, the condition for achieving correct prediction for the student SLM distilled from the teacher LLM's top-1 prediction coincides with the condition of the teacher LLM, i.e., $\bm{R}_{c,c}>\bm{R}_{c,c'}+\frac{\theta}{\phi-\theta},\forall c, c'\neq c$.

In the following paragraph, we justify when $\bm{R}_{c,c}>\bm{R}_{c,c'},\forall c, c'\neq c$, the student SLM distilled from the teacher LLM's top-2 prediction can achieve correct prediction. With Eq.(\ref{eq:p_condition}), we have when $\bm{R}_{c,c}>\bm{R}_{c,c'},\forall c, c'\neq c$, the teacher model's top-2 prediction always includes the true label $y_i$. Denote $\bar{y}_i$ as:
\begin{equation*}
    \bar{y}_i=\argmax_{c\neq y_i} \bm{R}_{y_i, c}
\end{equation*}
the training target $\bm{q}_i$ for distilling the teacher model's top-2 prediction can be expressed as:
\begin{equation}
    \begin{aligned}
        \bm{q}_i = \begin{cases}
        \frac{1}{2}\mathbf{e}(y_i) + \frac{1}{2}\mathbf{e}(\bar{y}_i), & \bm{x}_i \text{ is clean} \\
        \frac{1}{2}\mathbf{e}(y_i) + \frac{1}{2}\mathbf{e}(\tilde{y}_i), & \bm{x}_i \text{ is noisy} \\
    \end{cases}
    \end{aligned}
\end{equation}
Then, with the balance assumption, the second term in Eq.(\ref{eq:p_quantified}) is given as:
\begin{equation}
\begin{aligned}
    \frac{C}{m}\sum_{j:y_j=y_i}\bm{q}_j &= \frac{1}{2}\mathbf{e}(y_i) + \frac{1}{2}\bm{R}_{y_i, y_i}\mathbf{e}(\bar{y}_i) \\
    &\quad+ \frac{1}{2}\sum_{c\neq y_i} \bm{R}_{y_i, c}\mathbf{e}(c)
\end{aligned}
\end{equation}
Thus, if sample $\bm{x}_i$ is clean then $\left[\bm{p}_i\right]_c = $
\begin{equation}
    \begin{aligned}
        \begin{cases}
        \frac{\theta}{2} + \frac{\phi-\theta}{2} + \frac{1-\phi}{C}, &c = y_i; \\
        \frac{\theta}{2} + \frac{\phi-\theta}{2}\left(\bm{R}_{y_i, y_i} + \bm{R}_{y_i, \bar{y}_i}\right) + \frac{1-\phi}{C}, &c = \bar{y}_i; \\
        \frac{\phi-\theta}{2}\bm{R}_{y_i, c} + \frac{1-\phi}{C}, &c \neq y_i, \bar{y}_i. \\
    \end{cases}
    \end{aligned}
\end{equation}
where obliviously the max prediction is $y_i$ since $\sum_{c=1C}^C\bm{R}_{y_i, c}=1$. Then, if sample $\bm{x}_i$ is noisy and $\tilde{y}_i = \bar{y}_i$, $\left[\bm{p}_i\right]_c = $:
\begin{equation}
    \begin{aligned}
        \begin{cases}
        \frac{\theta}{2} + \frac{\phi-\theta}{2} + \frac{1-\phi}{C}, &c = y_i; \\
        \frac{\theta}{2} + \frac{\phi-\theta}{2}\left(\bm{R}_{y_i, y_i} + \bm{R}_{y_i, \tilde{y}_i}\right) + \frac{1-\phi}{C}, &c = \tilde{y}_i; \\
        \frac{\phi-\theta}{2}\bm{R}_{y_i, c} + \frac{1-\phi}{C}, &c \neq y_i, \bar{y}_i. \\
    \end{cases}
    \end{aligned}
\label{eq:p_noisy1}
\end{equation}
and when $\tilde{y}_i \neq \bar{y}_i$, $\left[\bm{p}_i\right]_c = $:
\begin{equation}
\label{eq:p_noisy2}
    \begin{aligned}
        \begin{cases}
        \frac{\theta}{2} + \frac{\phi-\theta}{2} + \frac{1-\phi}{C}, &c = y_i; \\
        \frac{\theta}{2} + \frac{\phi-\theta}{2}\bm{R}_{y_i, \tilde{y}_i} + \frac{1-\phi}{C}, &c = \tilde{y}_i; \\
        \frac{\phi-\theta}{2}\left(\bm{R}_{y_i, y_i} + \bm{R}_{y_i, \bar{y}_i}\right) + \frac{1-\phi}{C}, &c = \bar{y}_i. \\
        \frac{\phi-\theta}{2}\bm{R}_{y_i, c} + \frac{1-\phi}{C}, &c \neq y_i, \tilde{y}_i, \bar{y}_i. \\
    \end{cases}
    \end{aligned}
\end{equation}
Eq.(\ref{eq:p_noisy1}) and (\ref{eq:p_noisy2}) also yield $y_i$ as the max prediction of $\bm{p}_i$, which indicates the student SLM distilled from the teacher LLM's top-2 prediction achieves accurate predictions.

To sum up, the condition of achieving accurate prediction, i.e., achieving 100\% accuracy for either the pre-trained teacher LLM or the SLM distilled from the teacher LLM's top-1 prediction is:
\begin{equation}
    \label{eq:p_before1}
    \bm{R}_{c, c} > \bm{R}_{c, c'} + \frac{\theta}{\phi - \theta},~\forall c, c'\neq c
\end{equation}
and the condition of achieving 100\% accuracy for the SLM distilled from the teacher LLM's top-2 prediction is:
\begin{equation}
    \label{eq:p_before2}
    \bm{R}_{c, c} > \bm{R}_{c, c'},~\forall c, c'\neq c
\end{equation}
Since $\bm{R}_{c, c}$ reflects the clean probability, we replace $\bm{R}_{c, c}$ in Eq.(\ref{eq:p_before1}) and (\ref{eq:p_before2}) by $1-\sum_{i\neq c}\bm{R}_{c,i}$ that reflects the noise rates, which directly yields the conclusion in Eq.(\ref{eq:p_after1}) and (\ref{eq:p_after2}). These illustrate that the SLM distilled from LLM's top-2 prediction achieves 100\% accuracy with a more tolerant condition on label noise, providing the theoretical foundation of our proposed CanDist framework.

\begin{table*}[!t]
    \caption{Full prompts of prompting single (SA) and candidate ($\text{CA}_\text{add}$ and $\text{CA}_\text{all}$) annotations on the TREC dataset.}
    \label{tab:prompts_full}
    \centering
    \small
    \renewcommand\arraystretch{1.2}
    \scalebox{0.97}{
    \begin{tabular}{p{1.0cm}p{13.7cm}}
    \toprule
        \textbf{Strategy} & \textbf{Prompt} \\ 
    \midrule
        SA & \parbox{13.7cm}{\textsf{You are a helpful assistant for the task of question classification on the TREC (The Text REtrieval Conference Question Classification) dataset. You reply with brief, to-the-point answers with no elaboration as truthfully as possible. TREC dataset contains 5452 questions, each question is identified as one of the 6 types with respect to what it asks for: DESC; ENTY; ABBR; HUM; LOC; NUM, which stand for \textbf{Abbreviation}; \textbf{Description and abstract concepts}; \textbf{Entities}; \textbf{Human beings}; \textbf{Locations}; \textbf{Numeric values}, respectively. Each of these 6 classes contains a non-overlapping set of fine-grained sub-classes as follows: ABBR (Abbreviation): [Abbreviation and Expression abbreviated], DESC (Description and abstract concepts): [Definition of something. Description of something. Manner of an action and Reason.], ENTY (Entities):  [Animal. Organ of body; Color; Invention, book and other creative piece; Currency name; Disease and medicine; Event; Food; Musical instrument; Language; Letter like a-z; Other entity; Plant; Product; Religion; Sport; Element and substance. Symbols and sign. Techniques and method. Equivalent term. Vehicle. Word with a special property.], HUM (Human beings): [Group or organization of persons; Individual; Title of a person; Description of a person], LOC (Locations):  [City; Country; Mountain; Other location. State], NUM (Numeric values): [Postcode or other code; Number of something; Date; Distance, linear measure; Price; Order, rank; Other number; Lasting time of something; Percent, fraction; Speed; Temperature; Size, area and volume; Weight]. Your task is to classify the the given question as one of the 6 given coarse classes (ABBR, DESC, ENTY, HUM, LOC and NUM) based on what is asked and type of the answer. Given a question: $\dots$ What does this question ask about? Please identify the question \highlight{into one} of the six mentioned types.}} \\ 
    \midrule
        $\text{CA}_\text{add}$ & \parbox{13.7cm}{\textsf{You are a helpful assistant for the task of question classification on the TREC (The Text REtrieval Conference Question Classification) dataset. You reply with brief, to-the-point answers with no elaboration as truthfully as possible. TREC dataset contains 5452 questions, each question is identified as one of the 6 types with respect to what it asks for: DESC; ENTY; ABBR; HUM; LOC; NUM, which stand for \textbf{Abbreviation}; \textbf{Description and abstract concepts}; \textbf{Entities}; \textbf{Human beings}; \textbf{Locations}; \textbf{Numeric values}, respectively. Each of these 6 classes contains a non-overlapping set of fine-grained sub-classes as follows: ABBR (Abbreviation): [Abbreviation and Expression abbreviated], DESC (Description and abstract concepts): [Definition of something. Description of something. Manner of an action and Reason.], ENTY (Entities):  [Animal. Organ of body; Color; Invention, book and other creative piece; Currency name; Disease and medicine; Event; Food; Musical instrument; Language; Letter like a-z; Other entity; Plant; Product; Religion; Sport; Element and substance. Symbols and sign. Techniques and method. Equivalent term. Vehicle. Word with a special property.], HUM (Human beings): [Group or organization of persons; Individual; Title of a person; Description of a person], LOC (Locations):  [City; Country; Mountain; Other location. State], NUM (Numeric values): [Postcode or other code; Number of something; Date; Distance, linear measure; Price; Order, rank; Other number; Lasting time of something; Percent, fraction; Speed; Temperature; Size, area and volume; Weight]. Your task is to classify the the given question as one of the 6 given coarse classes (ABBR, DESC, ENTY, HUM, LOC and NUM) based on what is asked and type of the answer. Given a question: $\dots$ What does the question ask about? Please identify the question into one of the six mentioned types. \highlight{If you are unsure about your answer, please include other potential choices.}}} \\
    \midrule
        $\text{CA}_\text{all}$ &  \parbox{13.7cm}{\textsf{You are a helpful assistant for the task of question classification on the TREC (The Text REtrieval Conference Question Classification) dataset. You reply with brief, to-the-point answers with no elaboration as truthfully as possible. TREC dataset contains 5452 questions, each question is identified as one of the 6 types with respect to what it asks for: DESC; ENTY; ABBR; HUM; LOC; NUM, which stand for \textbf{Abbreviation}; \textbf{Description and abstract concepts}; \textbf{Entities}; \textbf{Human beings}; \textbf{Locations}; \textbf{Numeric values}, respectively. Each of these 6 classes contains a non-overlapping set of fine-grained sub-classes as follows: ABBR (Abbreviation): [Abbreviation and Expression abbreviated], DESC (Description and abstract concepts): [Definition of something. Description of something. Manner of an action and Reason.], ENTY (Entities):  [Animal. Organ of body; Color; Invention, book and other creative piece; Currency name; Disease and medicine; Event; Food; Musical instrument; Language; Letter like a-z; Other entity; Plant; Product; Religion; Sport; Element and substance. Symbols and sign. Techniques and method. Equivalent term. Vehicle. Word with a special property.], HUM (Human beings): [Group or organization of persons; Individual; Title of a person; Description of a person], LOC (Locations):  [City; Country; Mountain; Other location. State], NUM (Numeric values): [Postcode or other code; Number of something; Date; Distance, linear measure; Price; Order, rank; Other number; Lasting time of something; Percent, fraction; Speed; Temperature; Size, area and volume; Weight]. Your task is to classify the the given question as one of the 6 given coarse classes (ABBR, DESC, ENTY, HUM, LOC and NUM) based on what is asked and type of the answer. Given a question: $\dots$ What does the question ask about? Please identify the question \highlight{with all possible choices} of the six mentioned types.}} \\ 
    \bottomrule
    \end{tabular}}
\end{table*}

\section{Full Prompt Design}\label{subsec:full_prompts}
The full prompt designs of single annotations and candidate annotations are listed in Table \ref{tab:prompts_full}.

\end{document}